\documentclass[sigconf,nonacm]{aamas} 

\usepackage{balance} 
\usepackage{balance} 
\usepackage{booktabs}
\usepackage{amsmath}
\usepackage{graphicx}
\usepackage{tikz}
\usepackage{todonotes}
\usetikzlibrary{arrows.meta}
\usetikzlibrary{positioning}
\usetikzlibrary{calc}
\usepackage{algorithm}
\usepackage{algpseudocode}
\usepackage{placeins}

\newcommand{\NW}{\mathcal{NW}}
\newcommand{\alg}[1]{\textsc{#1}}

\setcopyright{none}
\acmConference{}{}{}{}
\acmBooktitle{}
\acmDOI{}
\acmISBN{}

\fancypagestyle{preprint}{%
  \fancyhf{}
  \fancyhead[L]{\footnotesize Preprint. Under review.}
  \fancyfoot[C]{\footnotesize\thepage}

}

\title[SSP]{Self-Referenced Social Preferences: Cooperation without Observing Others' Rewards}

\author{Mohamed A. Mohamed}
\affiliation{
  \institution{Amazon}
  \city{Vancouver}
  \country{Canada}}
\email{mmayman@amazon.com}

\author{Harshil Kotamreddy}
\affiliation{
  \institution{Nvidia}
  \city{Santa Clara}
  \country{United States}}
\email{hkotamreddy@nvidia.com}

\author{Marcos Jose}
\affiliation{
  \institution{Itaú Unibanco}
  \city{São Paulo}
  \country{Brazil}}
\email{marcos.jose@itau-unibanco.com.br}

\begin{abstract}
Social preferences can promote cooperation in multi-agent reinforcement learning, but existing approaches often require agents to observe the rewards of their peers. In many real-world interactions, however, an agent can, as humans do, observe others' behavior and outcomes without access to their private reward signals. We introduce self-referenced social preferences, in which each agent learns a model of its own reward, applies it to other agents' observed transitions to assess their outcomes from its own perspective, and feeds these self-referenced assessments into standard social preferences. We study two ways to incorporate these assessments: modifying the learning reward, or using them to weight policy updates. We evaluate the approach on three sequential social dilemmas, Escape Room, Clean Up, and Commons Harvest, which require volunteering, public-good contribution, and resource restraint, respectively. Across all three environments, agents learn cooperative behavior without observing others' rewards, including in settings where independent learners fail to cooperate, and frequently achieve more equitable divisions of jointly produced returns than agents with access to true rewards. The effective integration point depends on the social preference: inequity aversion works best in the reward together with a value look-ahead, whereas a purely benevolent preference benefits from policy-update weighting. Under partial observability, the policy-update approach continues to support cooperation. These results show that explicit access to other agents' reward signals is not necessary for learning cooperative behavior: social preferences can instead be grounded in self-referenced assessments of others' outcomes derived from their observed behavior.\looseness=-1
\end{abstract}


\newcommand{\BibTeX}{\rm B\kern-.05em{\sc i\kern-.025em b}\kern-.08em\TeX}

\begin{document}




\maketitle 
\pagestyle{preprint}
\thispagestyle{preprint}

\section{Introduction}
\label{sec:intro}

Cooperation in multi-agent reinforcement learning (MARL) often requires that agents learn about one another. The difficulty is particularly pronounced in \emph{sequential social dilemmas}~\cite{Leibo2017}, where individual incentives can conflict with collective welfare. In Escape Room, some agents must pull a costly lever so that others can leave; in Clean Up, agents must clean a river without direct reward so that apples continue to grow; and in Commons Harvest, agents must leave apples uneaten so that the resource can regenerate. Independent deep reinforcement learners fail to cooperate in these games~\cite{Perolat2017,Hughes2018,Yang2020}. A common remedy is to give each agent a \emph{social preference}, a term in its objective that depends on how the other agents are doing. Inequity-averse agents are penalized for earning more or less than their peers~\cite{Hughes2018}, agents with a social value orientation weight the others' outcomes alongside their own~\cite{McKee2020}, and prosocial agents add the others' rewards to theirs~\cite{Peysakhovich2018}. However, while each of these increase cooperation, they do so by reading the rewards of the other agents.

This is somewhat unsatisfying. A reward is an agent's private signal of how good an experience was, and giving one agent another's reward provides information that it could not obtain by interacting with the world. People do not see how rewarding an experience is for someone else. Autonomous vehicles from different manufacturers have no reason to report their objectives to one another. A prosocial agent that relies on the others' rewards therefore depends on a signal that exists in simulation but rarely outside it, and its success in simulation does not tell us whether cooperation can be learned without that signal. We relax this requirement by allowing agents to receive only the observations and actions of the other agents. Centralized-training methods already provide access to other agents' observations and actions during training~\cite{Lowe2017,Foerster2018COMA}; we retain this assumption while withholding their rewards, critics, and parameters.\looseness=-1

Humans and other animals face the same problem. They observe what others do, but not how rewarding it is for them, and they use \textit{empathy} to infer another's situation by interpreting observable cues through their own experience~\cite{Preston2002,Decety2004}. Psychology separates empathy into two components: \textit{cognitive} and \textit{motivational} empathy.  Cognitive empathy involves estimating how another is doing, while motivational empathy is the inner drive to care about their welfare and take action to help them ~\cite{Decety2004,deWaal2008}. Social preferences, which come from the study of human behaviour~\cite{Fehr1999,Fehr2000,Bolton2000,Charness2002,Henrich2001}, model motivational empathy. Existing multi-agent methods that use social preferences do not model cognitive empathy because they directly provide the agent a numeric value that tells them how another agent is doing. One way humans perform cognitive empathy is through simulation, imagining themselves in the other's situation~\cite{Gordon1986,Goldman2006}. We therefore ask: \textit{can agents in MARL learn cooperative behavior when they cannot directly use other agents' rewards within their social preferences, but can instead assess other agents through their observed behavior and outcomes and the agent's own experience?}\looseness=-1

To answer this question, we study MARL in settings where each agent has access only to its own reward, but where the reward for a given observation and action is the same for every agent. An agent can then learn a model of its own reward from its own transitions and apply it to the transitions of another agent. This allows the agent to use its own reward model to inform its social preferences. When reward functions are shared, it can recover the other agents' reward; when they differ, it instead evaluates the other agents' situation using the its own reward model. We call this \emph{self-reference}. The predicted rewards are given to an unchanged social preference, and each policy still acts only on its own observation. We explore two ways to use this prediction: \emph{Self-Referenced Reward} (SRR), adds the predicted reward to the agent's reward, and \emph{Self-Referenced Advantage} (SRA), adds the predicted reward to the coefficient of the policy update.\looseness=-1


In all three games, the best self-referenced learner exceeds plain PPO in both collective return and Nash welfare. In Escape Room and Clean Up it matches the agents that observe the true rewards. With inequity aversion in Clean Up, the self-referenced agents even produce more than the agents that observe the true rewards, and leave the worst-off agent more than twice as much. Commons Harvest is the exception: the agents that use true rewards give the highest collective return. It is the game where our estimates are weakest, because the reward model reads the policy's image features and is dominated by the large beam penalty, and the extra production goes almost entirely to one agent, so the true rewards raise production but not welfare.

A high collective return does not mean that every agent is well off. Someone has to pull the lever or clean the river, and they are not designated beforehand. We find that the most productive configurations often leave one agent with almost nothing, so we also report the lowest agent return and select configurations based on both Nash welfare and collective return. Because the same agents meet in every episode, we also give each agent a \emph{standing ledger} of everyone's predicted returns across episodes, which discounts the gains of an agent that has been ahead. With this change, the costly role in Escape Room is shared across agents instead of always falling on the same ones.

Our contributions are (i)  a MARL setting in which agents observe peers' transitions but not their reward signals; (ii) \alg{SRR}, \alg{SRA}, and a standing ledger, which use each agent's own rewards to infer peer outcomes; and (iii) an empirical study showing when self-referenced assessments recover the benefits of true peer rewards, when they do not, and how they affect productivity and fairness across different social preferences, sequential social dilemmas, partial observability, repeated interaction, and heterogeneous private valuations.\looseness=-1

\section{Background}
\label{sec:background}
In this section, we introduce the setting, the learning algorithm our agents use, and the social preferences we build on. We use subscripts for agents and superscripts for time: $r_i^t$ is the reward of agent $i$ at time $t$, $[z]_+ = \max(z, 0)$, and $\bar z_{-i} = \frac{1}{N-1}\sum_{j \ne i} z_j$ is the average of a quantity over the agents other than $i$.\looseness=-1
~\vspace{-5.0pt}
\subsection{Markov Games and Sequential Social Dilemmas}
\label{sec:background-games}

We consider $N$ agents in a general-sum, partially observable Markov game~\cite{Shapley1953,Littman1994,Hansen2004}. At each time step $t$, each agent $i$ receives an observation $o_i^t$, a function of the state of the environment, chooses an action $a_i^t$ according to its policy $\pi_i(a_i \mid o_i)$, and receives a reward $r_i^t$; the next state depends on the actions of all agents. Each agent maximizes its own expected discounted return, $\mathbb{E}\bigl[\sum_t \gamma^t r_i^t\bigr]$ with $\gamma \in [0, 1)$, and there is no team reward. In our main experiments every agent observes the full state, arranged around itself; Section~\ref{sec:results-po} restricts this view, which makes the games partially observable.

A \emph{sequential social dilemma} is a Markov game in which the policies that are individually best for each agent lead to lower returns for all than a cooperative joint policy that a single defector could exploit~\cite{Leibo2017}. In Escape Room, a public good is produced only if enough agents pay a private cost at the same time; in Clean Up and Commons Harvest agents share a renewable resource that individual consumption degrades, a common-pool resource problem~\cite{Perolat2017}.

\subsection{Actor-Critic Learning and PPO}
\label{sec:background-ppo}

Each agent $i$ improves its policy $\pi_i(a \mid o; \theta_i)$ by the policy gradient~\cite{Sutton2000,Williams1992}, which weights $\nabla \log \pi_i(a_i^t \mid o_i^t)$ by how much better things went after action $a_i^t$ than expected. Actor-critic methods implement this coefficient using a target function $\Psi_i^t$ that depends on the action $a_i$:
\begin{equation}
\left[
\nabla_{\theta} J(\theta)
=
\mathbb{E}_{\pi_{\theta}}
\left[
\sum_{0 \le t \le T}
\nabla_{\theta} \ln \pi_{\theta}(a_i^t \mid s_i^t)
\cdot
\Psi_i^t
\;\middle|\;
S_0 = s_0
\right]
\right]
\label{eq:target-fn}
\end{equation}

In PPO, $\Psi_i^t$ is the \emph{advantage} $A_i^t$, which is estimated using GAE~\cite{Schulman2016}. Any term in the target function $\Psi_i^t$ that does not depend on the action can be subtracted from the return without changing the gradient in expectation; it acts as a \emph{baseline}~\cite{Sutton2018}. Supplementary Appendix A.1 gives the policy-gradient, baseline, and GAE equations. Our agents use proximal policy optimization (PPO)~\cite{Schulman2017}, and each agent runs it independently~\cite{deWitt2020}.\looseness=-1

\subsection{Social Preferences}
\label{sec:background-social}

A social preference makes an agent's objective depend on how the other agents are doing, an idea from behavioural economics~\cite{Fehr1999} and social psychology~\cite{Messick1968}. Given to reinforcement learners as an intrinsic reward, such preferences change what independent learners converge to in sequential social dilemmas~\cite{Hughes2018,McKee2020}; Section~\ref{sec:related} reviews this work.\looseness=-1

In a sequential game a single reward says little about who is ahead, so
Hughes et al.~\cite{Hughes2018} measure how well agent $j$ is doing by a
discounted running sum of its rewards,
\begin{equation}
  e_j^t = \gamma\lambda_r\, e_j^{t-1} + r_j^t ,
  \label{eq:bg-trace}
\end{equation}
restarted at the start of every episode, and let each agent maximize the
return of a \emph{social utility}, its own reward plus a social term:
\begin{equation}
  u_i^t = r_i^t + F_i\bigl(e_1^t, \dots, e_N^t\bigr).
  \label{eq:problem-shaped}
\end{equation}
With $x_j = e_j^t$, we use three social terms:
\begin{align}
  F_i^{\mathrm{EI}}(x) &= \alpha_i\, \bar x_{-i},
  \qquad
  F_i^{\mathrm{SVO}}(x) = \alpha_i \bigl(\cos\phi_i\, x_i + \sin\phi_i\, \bar x_{-i}\bigr), \nonumber\\
  F_i^{\mathrm{IA}}(x) &= -\frac{\alpha_i}{N-1}\sum_{j \neq i}[x_j - x_i]_+
                         -\frac{\beta_i}{N-1}\sum_{j \neq i}[x_i - x_j]_+ .
  \label{eq:ssa-operators}
\end{align}
The three terms encode different reasons to care about others. EI
(empathetic influence) is pure benevolence: the agent gains when the
others' average outcome rises, as with the prosocial reward of
Peysakhovich and Lerer~\cite{Peysakhovich2018}. SVO fixes a ratio between
one's own outcome and the others' through an angle
$\phi_i$~\cite{Messick1968,McKee2020}: $\phi_i = 0$ is purely self-regarding
and $\phi_i = \pi/4$ weights both equally. IA is the inequity aversion of
Fehr and Schmidt~\cite{Fehr1999} and cares about differences rather than
levels: it penalizes being behind the others, with weight $\alpha_i$, and being ahead of them, with weight $\beta_i$. With IA, Eq.~\eqref{eq:problem-shaped} is the learner of Hughes et al. The two IA weights act differently in the
dilemmas we study: aversion to being ahead makes an agent share the work
that benefits everyone, as cleaning does in Clean Up, whereas aversion to
being behind gives it a reason to punish agents that take more, as by
firing a beam in Commons Harvest~\cite{Hughes2018}. Whatever the reason is,
each of these terms needs $e_j$ for every other agent $j$, and so the
other agents' rewards.

A social term can be implemented within PPO (and other actor-critic methods) in two places: in the reward $r_i$, which changes both the advantage and what the critic learns, or in the coefficient $A_i^t$ that weights each action, which changes the direction of the update and leaves the reward and the critic unchanged. The baseline property limits what a term can do in the coefficient: a term added to the coefficient changes the expected update only if it depends on the action, otherwise it acts as a baseline. Our two methods modify the update in these two places respectively. Neither is specific to PPO: every reinforcement learning algorithm learns from a reward, and every actor-critic weights its policy gradient by an advantage. We use PPO as a representative actor-critic learner because it is widely used in recent work on sequential social dilemmas~\cite{deWitt2020,Leibo2021}.\looseness=-1
\section{Problem Formulation}
\label{sec:problem}

The social utility of Eq.~\eqref{eq:problem-shaped} requires every agent to know the rewards of all the other agents. Here we define the setting in which an agent knows only its own reward, and show how empathy can be modeled in such a setting.\looseness=-1

\paragraph{Withholding reward signals.}
\label{sec:ssa-information}
We consider the Markov game of Section~\ref{sec:background-games} with $N$ independent learners. During training, every agent $i$ receives the transitions of all agents, with their episode boundaries, but only its own reward:
\begin{equation}
  \mathcal{D}_i = \bigl\{(o_j^t, a_j^t, o_j^{t+1}) : j = 1, \dots, N,\ t \ge 0\bigr\}
  \cup \bigl\{r_i^t : t \ge 0\bigr\}.
  \label{eq:problem-data}
\end{equation}
Methods with centralized training and decentralized execution~\cite{Lowe2017,Foerster2018COMA} and methods that model other agents~\cite{Raileanu2018,Rabinowitz2018,Jaques2019} already give agents the others' observations and actions; the difference here is that the others' rewards are withheld. The agents share no parameters, critics, or reward models and exchange no messages, and at execution each policy $\pi_i(a_i \mid o_i)$ acts on its own observation.

\paragraph{Modeling empathy.}
From $\mathcal{D}_i$, agent $i$ can compute its own trace $e_i$ but not the traces $e_j$ of the others, and so it cannot compute $F_i$. It therefore faces two problems. It must estimate, from $\mathcal{D}_i$ alone, a quantity $x_{ij}$ that can stand in for $e_j$, which is the cognitive part of empathy, and it must decide how to use that estimate when it learns, which is the motivational part. An agent that is given the rewards $r_j$ computes $e_j$ exactly, and we use it as the true-reward reference.\looseness=-1

\paragraph{Estimating the rewards of other agents.}
Self-reference is most direct when every agent's reward comes from the same reward function $R$,
\begin{equation}
  r_j^t = R(o_j^t, a_j^t, o_j^{t+1}) \qquad \text{for all } j,
  \label{eq:problem-assumption}
\end{equation}
where each observation is arranged relative to its own agent, so that $R$ applies from every agent's position. Under this condition, a model of one's own reward can approximate the other's actual reward. The condition holds in our main experiments: a reward comes from an event in the agent's own transition, such as eating an apple, pulling the lever, reaching the open door, or firing or being hit by a beam, and every agent is rewarded for these events in the same way. Agent $i$ can then learn $R$ from its own transitions and apply it to $j$'s, provided that $j$'s transition shows the rewarded event and resembles transitions that $i$ has experienced. Two caveats remain. (1) When a transition does not fully determine the reward, a model fitted by regression estimates its mean, and a nonlinear social term computed from the mean, such as inequity aversion with its $[\cdot]_+$, need not equal the same term computed from the actual rewards. (2) When the reward functions differ, the estimate is no longer $j$'s reward but the reward that $i$ would receive in $j$'s place; Section~\ref{sec:results-hetero} tests this case.\looseness=-1~\vspace{-5pt}

\section{Self-Referenced Social Preferences}
\label{sec:ssa}

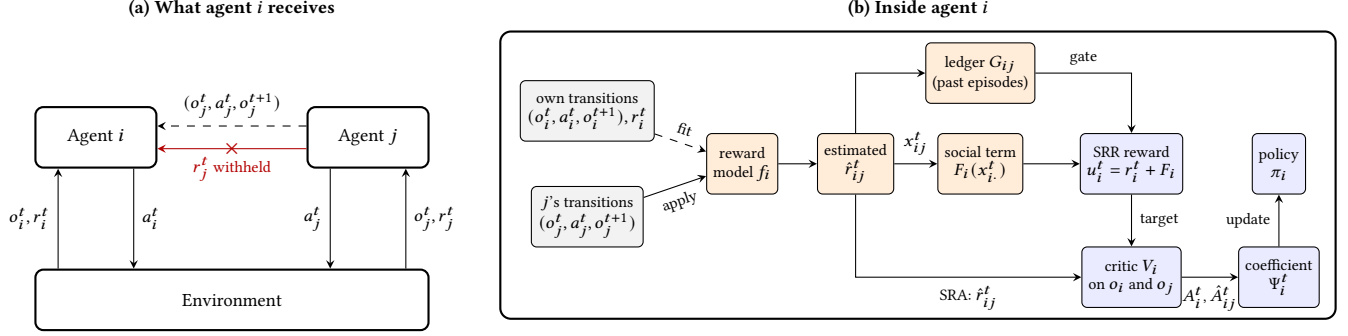
\begin{figure*}[t]
  \centering
  \vspace{-15pt}
  \begin{tikzpicture}[
      font=\scriptsize, >=stealth,
      agent/.style={draw, thick, rounded corners=3pt, align=center,
                    minimum width=16mm, minimum height=8mm, font=\footnotesize},
      comp/.style={draw, rounded corners=2pt, align=center, inner sep=2.5pt,
                   minimum height=8mm, fill=orange!15},
      std/.style={comp, fill=blue!8},
      dat/.style={comp, fill=black!5},
      lab/.style={font=\scriptsize, align=center}]
    \begin{scope}
      \node[agent] (Ai) at (0, 2.6) {Agent $i$};
      \node[agent] (Aj) at (3.6, 2.6) {Agent $j$};
      \node[agent, minimum width=52mm] (Env) at (1.8, 0.45) {Environment};
      \draw[->] ([xshift=-5mm]Env.north -| Ai.south) -- node[left, lab]{$o_i^t, r_i^t$} ([xshift=-5mm]Ai.south);
      \draw[->] ([xshift=5mm]Ai.south) -- node[right, lab]{$a_i^t$} ([xshift=5mm]Ai.south |- Env.north);
      \draw[->] ([xshift=5mm]Env.north -| Aj.south) -- node[right, lab]{$o_j^t, r_j^t$} ([xshift=5mm]Aj.south);
      \draw[->] ([xshift=-5mm]Aj.south) -- node[left, lab]{$a_j^t$} ([xshift=-5mm]Aj.south |- Env.north);
      \draw[->, dashed] ([yshift=1.5mm]Aj.west) -- node[above, lab]{$(o_j^t, a_j^t, o_j^{t+1})$} ([yshift=1.5mm]Ai.east);
      \draw[->, red!70!black] ([yshift=-1.5mm]Aj.west) -- node[below, lab, text=red!70!black]{$r_j^t$ withheld} ([yshift=-1.5mm]Ai.east);
      \node[red!70!black, font=\small] at (1.8, 2.45) {$\times$};
      \node[font=\footnotesize\bfseries, anchor=south] at (1.8, 4.08) {(a) What agent $i$ receives};
    \end{scope}
    \begin{scope}[xshift=5.55cm]
      \node[dat]  (own)  at (0.95, 2.95) {own transitions\\$(o_i^t, a_i^t, o_i^{t+1}), r_i^t$};
      \node[dat]  (oth)  at (0.95, 1.55) {$j$'s transitions\\$(o_j^t, a_j^t, o_j^{t+1})$};
      \node[comp] (f)    at (3.0, 2.25) {reward\\model $f_i$};
      \node[comp] (rhat) at (4.5, 2.25) {estimated\\$\hat r_{ij}^t$};
      \node[comp] (F)    at (6.15, 2.25) {social term\\$F_i(x_{i\cdot}^t)$};
      \node[comp] (G)    at (6.15, 3.45) {ledger $G_{ij}$\\(past episodes)};
      \node[std]  (u)    at (8.15, 2.25) {SRR reward\\$u_i^t = r_i^t + F_i$};
      \node[std]  (V)    at (8.15, 0.75) {critic $V_i$\\on $o_i$ and $o_j$};
      \node[std]  (Psi)  at (10.1, 0.75) {coefficient\\$\Psi_i^t$};
      \node[std]  (pi)   at (10.1, 2.25) {policy\\$\pi_i$};
      \draw[->, dashed] (own) -- node[above, lab, sloped]{fit} (f);
      \draw[->] (oth) -- node[below, lab, sloped]{apply} (f);
      \draw[->] (f) -- (rhat);
      \draw[->] (rhat) -- node[above, lab]{$x_{ij}^t$} (F);
      \draw[->] (F) -- (u);
      \draw[->] (rhat) |- (G);
      \draw[->] (G) -| node[above, lab, pos=0.25]{gate} (u);
      \draw[->] (u) -- node[right, lab]{target} (V);
      \draw[->] (rhat) |- node[below, lab, pos=0.75]{SRA: $\hat r_{ij}^t$} (V);
      \draw[->] (V) -- node[below, lab]{$A_i^t$, $\hat A_{ij}^t$} (Psi);
      \draw[->] (Psi) -- node[left, lab]{update} (pi);
      \draw[thick, rounded corners=4pt] (-0.2, 0.2) rectangle (10.85, 4.0);
      \node[font=\footnotesize\bfseries, anchor=south] at (5.33, 4.08) {(b) Inside agent $i$};
    \end{scope}
  \end{tikzpicture}
    \caption{Self-referenced social preferences. (a) During training agent $i$ receives the transitions of every other agent $j$, but never $j$'s reward. (b) Agent $i$'s reward model, fitted on its own transitions, assesses $j$'s transitions, giving self-referenced estimates $\hat r_{ij}^t$, whose running sums $x_{ij}^t$ feed an unchanged social term $F_i$. \alg{SRR} adds the term to the reward; \alg{SRA} evaluates the critic on $j$'s observations and weights the scores into the policy-update coefficient $\Psi_i^t$. The ledger discounts $i$'s gains when $i$ has been ahead.\looseness=-1}
    \Description{Diagram illustrating self-referenced social preferences, agent observations, and reward modeling.}
    \vspace{-14pt}
\label{fig:method}
\end{figure*}

Agent $i$ cannot observe $r_j$, but it observes $j$'s transitions and knows its own rewards. It therefore learns a model of its own reward and applies it to $j$'s transitions, asking what it would have earned in $j$'s place. The key idea is not to reconstruct another agent's private reward exactly, but to evaluate its observed situation using the focal agent's own learned reward model. Figure~\ref{fig:method} shows the resulting flow of information inside one agent. These estimates are used in a social term that modifies the update rule in the two ways described in Section~\ref{sec:background-social}: modifying the reward, which we call \emph{Self-Referenced Reward}, and modifying the coefficient of the policy update, which we call \emph{Self-Referenced Advantage}. 
Supplementary Appendix A.2 gives one training iteration in Algorithm S1.\looseness=-1

\subsection{Self-Referenced Reward Estimates}
\label{sec:ssa-rewards}
Agent $i$ fits $f_i(o,a,o')$ by regression on its own transitions and
applies the same model to every observed transition,
\begin{equation}
  \mathcal{L}_{f_i}
  =\tfrac12\,\mathbb{E}_{\mathcal B_i}
  \bigl[\bigl(f_i(o_i^t,a_i^t,o_i^{t+1})-r_i^t\bigr)^2\bigr],
  \quad
  \hat r_{ij}^t=f_i(o_j^t,a_j^t,o_j^{t+1}),
  \label{eq:ssa-rm}
\end{equation}
where $\mathcal{B}_i$ is a minibatch of $i$'s own transitions, and with
$\hat r_{ii}^t:=r_i^t$. In practice, $f$ is learned using a shared representation with the policy.\looseness=-1

We call $\hat r_{ij}$ a self-referenced reward estimate. If the agents share a reward function, it estimates $j$'s reward; otherwise, it estimates the reward that $i$ would assign to $j$'s transition. Rare rewards create a coverage problem: an agent that only cleans may stop seeing apple rewards. We therefore add a second term to the loss of Eq.~\eqref{eq:ssa-rm} over a minibatch drawn from a replay of the agent's own nonzero-reward transitions. The social term is off for the first $W$ environment steps, while the model is fit. The estimates are treated as constants, so no gradient flows from the social term into the reward model.

\subsection{Estimating Future Outcomes}
\label{sec:ssa-operators}
Agent $i$ keeps one trace for every agent $j$, computed from its estimates exactly as Eq.~\eqref{eq:bg-trace} computes it from true rewards:
\begin{equation}
  e_{ij}^t=\gamma\lambda_r(1-s_j^t)e_{ij}^{t-1}+\hat r_{ij}^t,
  \label{eq:ssa-level}
\end{equation}
where $s_j^t$ marks the start of an episode. A trace judges an agent only by what it has already earned, so an agent that has just cleaned the river looks worse than the other agents, even though apples will soon grow for it. To amend this, the $+V$ variant adds a term that estimates what agent $i$ expects $j$ to earn from in the future. This is done using its critic $V_i^{\rm obj}$, estimating the return of $i$'s own objective, evaluated on $j$'s next observation:\looseness=-1
\begin{equation}
  x_{ij}^t=e_{ij}^t+w_V\gamma(1-d_{j,t}^{\rm term})V_i^{\rm obj}(o_j^{t+1}),
  \label{eq:ssa-level-v}
\end{equation}
where $d_{j,t}^{\rm term}$ marks the end of $j$'s episode and $w_V=(1-\gamma)/(1-\gamma\lambda_r)$ puts the value on the scale of the trace. The look-ahead is $i$'s value of being in $j$'s position, not an estimate of $j$'s reward. $x_{i\cdot}$ takes the place of the true traces in the social terms of Eq.~\eqref{eq:ssa-operators}.\looseness=-1~\vspace{-3pt}

\subsection{Self-Referenced Reward (SRR)}
\label{sec:ssa-reward-slot}

\alg{SRR} adds the social term of the self-reference to the agent's reward,\looseness=-1   
\begin{equation}
  u_i^t=r_i^t+F_i(x_{i\cdot}^t),
  \label{eq:ssa-reward-slot}
\end{equation}
and PPO computes the advantage and the critic target from $u_i$. With the true rewards in place of $\hat r_{ij}$, the same construction gives the true-reward learners, which we denote \alg{TR}; \alg{TR-IA} without $+V$ is the inequity-averse agent of Hughes et al.~\cite{Hughes2018}. Using reward estimates in place of the true reward brings one caveat: a poor estimate affects both the policy and the critic.\looseness=-1

\subsection{Self-Referenced Advantage (SRA)}
\label{sec:ssa-estimation}

\alg{SRA} instead uses the estimates in the policy update coefficient (see Eq.~\eqref{eq:target-fn}). It leaves $r_i$ and $V_i^{\rm obj}$ unchanged and changes only the coefficient that weights each action. Agent $i$ evaluates its critic on $j$'s observations and forms a temporal-difference error and its generalized advantage estimate, which we call the cross score,
\begin{align}
  \hat\delta_{ij}^t
  &=\hat r_{ij}^t-V_i^{\rm obj}(o_j^t)
    +\gamma(1-d_{j,t}^{\rm term})V_i^{\rm obj}(o_j^{t+1}),
  \label{eq:ssa-td}\\
  \hat A_{ij}^t
  &=\hat\delta_{ij}^t
    +\gamma\lambda_{\rm SRA}(1-d_{j,t}^{\rm end})\hat A_{ij}^{t+1},
  \label{eq:ssa-gae}
\end{align}
where $d^{\rm end}$ also marks a time-limit truncation. The score measures how much better $j$'s transition went than $i$'s critic expected, judged by $i$'s own reward model and values; it is not the advantage of either agent's policy, as the critic was trained on $i$'s observations and objective, not $j$'s.\looseness=-1

Let $q_{ii}$ be $i$'s own GAE and $q_{ij}$ the cross score of Eq.~\eqref{eq:ssa-gae}, each standardized over time and parallel environments. EI and SVO combine them linearly:
\begin{equation}
  \textstyle
  \Psi_{i,\mathrm{EI}}^t=q_{ii}^t+\alpha_i\bar q_{i,-i}^t\; \qquad
  \Psi_{i,\mathrm{SVO}}^t=(1+\alpha_i\cos\phi_i)q_{ii}^t+\alpha_i\sin\phi_i\bar q_{i,-i}^t.
  \label{eq:ssa-linear-coefficients}
\end{equation}
The 1 in front of $\alpha_i\cos\phi_i$ keeps the agent's own score in the update for every angle $\phi_i$. For IA, the levels decide the weights. For $j\ne i$,
\begin{align}
  w_{ij}^t
  &=\frac{\beta_i\mathbf 1[x_{ii}^t>x_{ij}^t]
          -\alpha_i\mathbf 1[x_{ii}^t<x_{ij}^t]}{N-1},
  \qquad
  w_{ii}^t=1-\sum_{j\ne i}w_{ij}^t,
  \nonumber\\
  \Psi_{i,\mathrm{IA}}^t
  &=w_{ii}^tq_{ii}^t+\sum_{j\ne i}w_{ij}^tq_{ij}^t.
  \label{eq:ssa-ia-coefficient}
\end{align}
Being ahead of $j$ moves weight from $i$'s own score to $j$'s; being behind $j$ puts negative weight on $j$'s score and adds the same weight to $i$'s own. The coefficient, normalized within each minibatch, is used as the target function in Eq.~\eqref{eq:target-fn}, and no gradient flows through it or into agent $j$. \alg{SRA} is therefore a cross-agent credit-assignment rule rather than an unbiased policy-gradient estimator of a fixed social objective. EI and SVO thus add up the agents' outcomes much as in the reward, whereas for IA the comparison only decides how the scores are weighted: nothing in the agent's return penalizes being ahead or behind, as it does in \alg{SRR}.

\section{Experiments}
\label{sec:experiments}

\subsection{Experimental Setup}
\label{sec:setup}

We evaluate our methods in three sequential social dilemmas. In Section~\ref{sec:results} we seek to answer the following questions: (1) Do agents cooperate without observing each other's rewards, (2) where should the self-referenced estimate be used in the learning update, (3) which agents take a penalty for cooperation to occur, and (4) whether self-referenced estimates are still useful when the environment is partially observable or rewards are heterogeneous .\looseness=-1

\subsubsection{The Three Games}
\label{sec:setup-envs}

The games require different behavior for cooperation to occur: volunteering, labor, and restraint. In the main experiments, all three games are fully observable, so the only thing an agent lacks is the other agents' rewards. Every observation is arranged around its own agent, which is what lets an agent apply its own reward model to another agent's observation; Supplementary Table S2 lists what each agent observes and can do.


\paragraph{Escape Room.}
ER($N$,$M$) is the Escape Room environment introduced by Yang et al.~\cite{Yang2020}. At every step, each of $N$ agents chooses \emph{stay}, \emph{lever}, or \emph{door}. Pulling the lever results in ~$-1$ reward; if at least $M = \lceil N/2 \rceil$ agents pull the lever in the same step, the door opens and every agent at the door earns $+10$. Episodes last at most five steps. The group optimum is $10(N-M)-M$ per episode (ex. 17 with $5$ agents). Selfish learners earn 0 reward since pulling the lever only ever pays somebody else. The optimum concentrates the cost on the $M$ agents at the lever unless they take turns over many episodes.\looseness=-1

\paragraph{Clean Up.}
Clean Up~\cite{Hughes2018} is a gridworld public-goods game: agents earn
$+1$ per apple eaten, apples stop growing when waste fills the river, and
cleaning the river restores growth but pays the cleaner nothing. We use a $25 \times 18$ map with 1{,}000-step episodes. Cooperation requires that some agents clean while the others eat, so the work and the apples can be divided very unevenly. The main results use five agents; Supplementary Tables S9 and S10 report 3 and 7 agents respectively.\looseness=-1

\paragraph{Commons Harvest.}
Commons Harvest~\cite{Perolat2017} is a common-pool resource game on a
$16 \times 38$ map with seven agents and 1{,}000-step episodes. Eating an
apple is worth $+1$. Apples regrow in empty cells with a probability $p$ at each time step. $p$ is 0 if there are no apples around the cell and p is larger when there are more apples around it. Thus, eating
the last apples of a patch destroys future apples for everyone. Each agent has a beam that allows it to give other agents penalties. The agent that shoots the beam receives a reward of -1 and the agent it hits receives -50.\looseness=-1

\subsubsection{The Learning Agents}
\label{sec:setup-agents}
Every agent is an independent PPO learner with its own actor, critic and, for self-referenced agents, reward model; nothing is shared between agents. We train agents in Escape room for 500k environment steps and agents in the Clean Up and Commons Harvest environments for 5M steps. In the partially observable setting, we train agents for 7M steps. Supplementary Appendix B.2 lists the networks, warm-up, and PPO settings.

We compare four families of learners. Two are baselines: \emph{plain PPO}, which has no social term, and \emph{the true-reward learners}, \alg{TR}, which add the operators of Eq.~\eqref{eq:ssa-operators}, computed from traces of the other agents' true rewards, to the reward; \alg{TR-IA} is the agent of Hughes et al.~\cite{Hughes2018}. The true-reward learners act as a comparison to show which settings benefit from providing agents with access to the true reward. \alg{SRR} adds the same operators, computed from self-referenced estimates, to the reward, with or without the look-ahead \alg{+V}. Without the look-ahead it differs from \alg{TR} only in the source of the rewards and in the warm-up of its reward model. The main figures show the reward learner with the look-ahead, \alg{SRR+V}; Section~\ref{sec:results-placement} compares both. \alg{SRA} puts the estimates in the coefficient of the policy update and leaves the reward and the critic unchanged.\looseness=-1

Each family comes with the EI, SVO, and IA operators, with weights tuned
for each game and group size, and we name a learner by family and
operator, as in \alg{SRR-IA}. A \emph{self-scaling control}, SVO with
$\phi = 0$, uses no information about the others; it tests whether a gain comes from the social term or merely from rescaling the agent's own reward. For the repeated game, we add a standing gate (see Supplementary Appendix) to four selected Escape Room configurations, \alg{SRA-EI} with three and five agents and \alg{SRR-IA} with five and seven, and change nothing else. For Section~\ref{sec:results-hetero}, two of the five Clean Up agents value an apple at 2 instead of 1; we reuse the settings selected under shared rewards and train on the same held-out seeds.

\paragraph{Selection and evaluation.}
\label{sec:setup-protocol}
We tune on three seeds in every game, and on two for the partial-view retuning, and score every candidate by the mean minus the standard deviation of a metric across the tuning seeds. We choose once by collective return and once by Nash welfare (see Section \ref{sec:setup-metrics}) and call these the learner's collective-return and Nash welfare selections. Unless stated otherwise, figures show the collective-return selection. Each choice is frozen and trained from scratch on eight seeds that played no part in tuning, and we never reselect on these runs. Intervals are two-sided Student-$t$ 95\% intervals across the eight seeds.\looseness=-1
\paragraph{Partial view.}
\label{sec:setup-po}
With the partial view, each agent in Clean Up and Commons Harvest sees only the $15\times15$ window around itself. The transition agent $i$ receives from $j$ then shows only $j$'s neighborhood, so $i$ must judge $j$'s situation from a view it may never have had itself. We retune every learner for this setting, with feed-forward agents and with agents whose policy and critic each carry a GRU~\cite{Cho2014}. The grid and the recurrent architecture can be found in Appendix B.4.\looseness=-1
\begin{figure*}[t]
  \centering
  \includegraphics[width=0.9\textwidth]{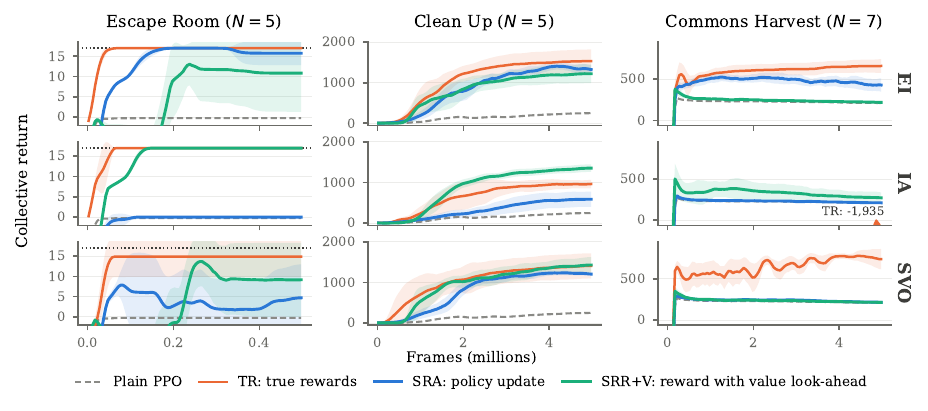}
    \vspace{-15pt}
    \caption{Collective return over training. Each panel shows plain PPO, the TR learner, and both of our methods: \alg{SRA} and \alg{SRR+V}. Lines are means over eight seeds and shaded regions are pointwise 95\% $t$ intervals. The dotted line is the Escape Room optimum, $17$. Commons Harvest \alg{TR-IA} is outside the plotted range, with final mean $C=-1{,}935$, and is marked at the panel edge.\looseness=-1}
    \vspace{-15pt}
  \label{fig:cr-curves}
\end{figure*}
\subsubsection{Outcome and Behavioral Measures}
\label{sec:setup-metrics}
Because every agent keeps the same index in every episode of a run, we can average each agent's return over many episodes. Let $\bar R_{i,s}$ be agent $i$'s episodic return averaged over the final 20{,}000 environment steps of seed $s$ in Clean Up and Commons Harvest and over the final window of training in Escape Room. From these $N$ averages we compute
\begin{equation}
  \textstyle
  C_s=\sum_{i}\bar R_{i},\;
  L_s=\min_i \bar R_{i},\;
  \NW_{c,s}=e^{\frac{1}{N}\sum_{i}\log(1+[\bar R_{i}+c]_+)}-1.
  \label{eq:collective-lowest}
\end{equation}
The collective return $C_s$ is how much the group produces, whoever receives it; the lowest agent return $L_s$ shows whether some agent is left behind; and the Nash welfare $\NW_{c,s}$~\cite{Kaneko1979,Caragiannis2019}, a geometric mean that for a fixed total is largest when the returns are equal, rewards both production and balance, with a shift $c$ of 5 in Escape Room and 0 elsewhere. Because returns are averaged over many episodes, these measures can tell agents that take turns at a costly role from agents whose roles stay fixed. In Escape Room, we also report each agent's \emph{exit rate}, the fraction of episodes in which it reaches the open door, and call the lever role shared on a seed when every agent has a positive return and the exit rates have a standard deviation below 0.20. In Commons Harvest, we report sustainability~\cite{Perolat2017}, the mean time step at which apples are eaten; higher means agents leave apples for later.\looseness=-1

\subsection{Results}
\label{sec:results}

\subsubsection{Can Agents Cooperate without Observing Others' Rewards?}
\label{sec:results-cooperate}

In all three games, social preferences can operate on self-referenced assessments without observing the others' rewards. How closely those assessments match the true rewards varies by game, but exact reward recovery is not necessary for cooperation. Figure~\ref{fig:cr-curves} compares plain PPO, the true-reward learners, and both of our methods for every game and operator.

Escape Room is the most straightforward test, because pulling the lever only ever benefits someone else, and plain PPO agents never learn to pull it. A self-referenced agent can still see the benefit: it has learned from its own exits that reaching the open door pays ten, so when another agent stands at the door, it recognises that reward. Thus, \alg{SRA-EI} and \alg{SRR-IA+V} reach the optimum, as the agents given the true rewards do. SVO on the other hand weighs the agent's own outcome as well as the others'. Thus, the benefit to other agents must outweigh the agent's own cost of pulling the lever. With true rewards and a strong weight it does, and the agents learn to pull. \alg{SRR-SVO} was selected with a much weaker weight, because that weight was already enough on the tuning seeds; with it, pulling still feels like a loss, and cooperation rests on the look-ahead alone, so it reaches only about half the optimum. \alg{SRA-SVO} caps how much weight it can give to the others (see Section~\ref{sec:results-placement}). Supplementary Appendix C.4 shows that at least one self-referenced learner reaches the optimum from two to twelve agents.\looseness=-1

Clean Up requires agents to sacrifice reward to provide labor. Cleaning the river earns nothing directly, and plain PPO agents rarely do it. Every self-referenced learner produces about as much as its true-reward counterpart, so self-referenced estimates are enough to sustain a public good. Under inequity aversion, SRR even does better than TR: \alg{SRR-IA+V} produces more than \alg{TR-IA} and more than doubles the return of its poorest agent. We think the true rewards are the obstacle. An agent that has just eaten an apple seems far ahead of one that is cleaning, although apples are about to grow for the cleaner. With a strong penalty on being ahead, agents given the true rewards stop eating, and \alg{TR-IA} harvests nothing on any tuning seed, so its selection settles on a weak penalty. \looseness=-1

Commons Harvest requires agents to restrain themselves. Agents must leave apples uneaten so that patches regrow, and an agent that eats the last apples of a patch takes future apples from everyone. \alg{SRA-EI} learns this restraint without seeing the others' rewards. It doubles plain PPO's collective return, and its agents eat apples four times later on average. This is due to the policy update. When an agent strips a patch, its critic values the other agents' positions less turning their cross scores turn negative. Thus, the action that stripped the patch loses credit. This is also the game where our self-referenced learners fall furthest behind the true rewards, and \alg{TR-EI} and \alg{TR-SVO} produce more still. Estimation is hardest here, as the reward model is learned through a shared representation with the policy, whose image features were not trained to recognize rewards, and the beam's large penalty dominates what it learns. Inequity aversion goes the other way. An agent averse to being behind has a reason to fire at an agent that is ahead, and with true rewards that gap is visible at every step, so the agents of \alg{TR-IA} almost stop harvesting or fire at one another~\cite{Hughes2018}. \alg{SRR-IA+V} was selected with a weaker penalty on being behind, so it has little reason to fire, and it keeps harvesting\looseness=-1.

Taken together, self-referenced assessments are enough to start and sustain cooperation in all games. But, in Commons Harvest, the true-reward learners produce more. In \ref{sec:results-who-pays}, we look at the distribution of return among agents when using each of these methods.\looseness=-1

\begin{figure*}[t]
  \centering
  \includegraphics[width=0.9\textwidth]{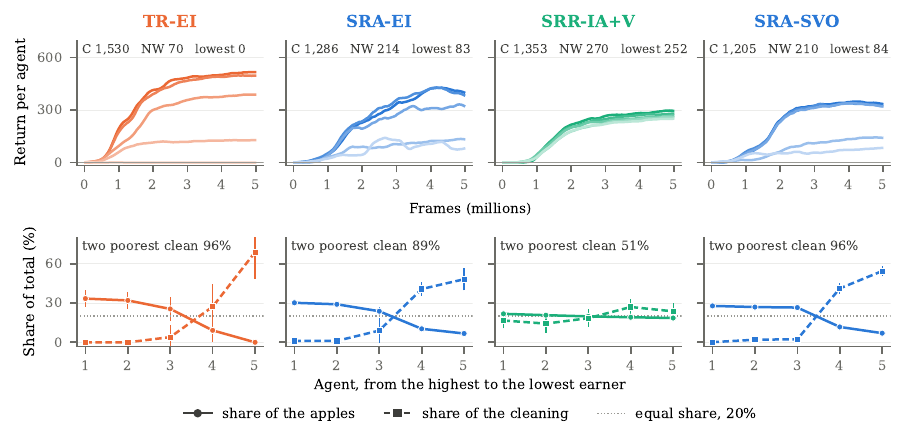}
  \vspace{-16pt}
  \Description{}
    \caption{Who receives the return and who does the work in Clean Up with five agents. Top: every agent's return over training; agents are ranked within each seed by their final return and averaged by rank over the eight seeds, darkest the richest. Bottom: each agent's final share of the apples, solid, and of the cleaning, dashed, with 95\% confidence intervals; 20\% is an equal share.}
      \vspace{-16pt}
  \label{fig:agents-cu}
\end{figure*}
\subsubsection{Where Should the Estimate Enter?}
\label{sec:results-placement}
\vspace{-8pt}
The two methods of self-reference suit different kinds of social preference. Inequity aversion works only with SRR: \alg{SRR-IA+V} is the better of our methods in every game, while \alg{SRA-IA} never escapes Escape Room. In the reward, being ahead costs the agent part of its return, which it can learn to avoid. In the policy update, being ahead only shifts weight between scores, (see Eq.~\eqref{eq:ssa-ia-coefficient}). Thus, nothing pushes it to stop being ahead. EI only adds up the others' outcomes, which credits actions with the gains of other agents. Thus, \alg{SRA-EI} is the better method in Escape Room and Commons Harvest and about as good in Clean Up. SVO also adds outcomes but weighs the agent's own as well. In the policy update this caps the weight assigned to the other agents at $\tan\phi_i$ times the agent's own weight, which is not enough to achieve the performance of EI.\looseness=-1
Thus, we note that a preference comparing agents belongs in the reward. A benevolent pereference such as EI is safer in the policy update.

The look-ahead is what makes SRR work. Without it, agents that never reach the door never see a reward to assess, so the social term stays at zero. The critic's value distinguishes positions before any reward is assessed and gets cooperation started. Under inequity aversion, it also prevents the agent misjudging the cleaner.Under EI, the look-ahead brings no clear gain outside Escape Room: in Clean up, SRR-EI+V produces about 290 apples less than \alg{SRR-EI}, and in Commons Harvest, it stays at plain PPO's level, far below \alg{SRR-EI}.\looseness=-1

Group size exposes the limit of SRA. Its coefficient averages over the $N-1$ others, so any one agent's action matters less as the group grows, and \alg{SRA-EI} falls further behind \alg{TR-EI}. SRR, on the other hand, does not dilute, and here self-reference again holds up better TR: IA with true rewards declines as the group size grows, while with self-referenced estimates and the look-ahead it holds. A larger group gives an inequity-averse agent more pairs to be behind in, and true rewards make each comparison sharper than it is. Appendix D.2 gives the results with $2-12$ agents.\looseness=-1\vspace{-4pt}

\begin{figure*}[t]
  \centering
  \includegraphics[width=0.88\textwidth]{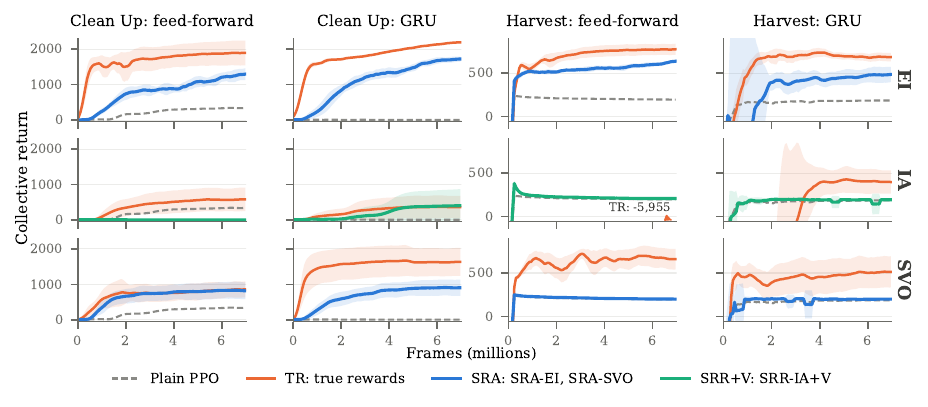}
  \vspace{-16pt}
    \caption{Partial view: collective return over training in Clean Up and Commons Harvest with a $15\times15$ window, for feed-forward and GRU agents. Lines are means over eight held-out seeds and shaded regions pointwise 95\% Student-$t$ intervals.}
      \vspace{-12pt}
  \label{fig:partial-obs}
\end{figure*}

\subsubsection{Who Pays for Cooperation?}
\label{sec:results-who-pays}
Collective return on its own does not give us the full picture. Figure~\ref{fig:agents-cu} shows which agents gained the most return and which agents cleaned the most in Clean Up. It shows the most productive TR learner and self-referenced learner for each operator; further details can be found in Appendix E.1.\looseness=-1

The learner that produces the most shares it worst. \alg{TR-EI} produces 1{,}530 apples, but its poorest agent earns nothing for the whole run, and the bottom row shows why: the richest agent eats a third of the apples and cleans almost nothing, while the two poorest do 96\% of the cleaning. The group has split into eaters and cleaners, and collective return counts both the same. Nash welfare does not: as a geometric mean, it equals the average return when every agent earns the same and falls towards zero when any agent is left with nothing, so \alg{TR-EI}'s is only 70, although its agents average about 300 apples. \alg{SRA-EI} and \alg{SRA-SVO} soften the split, raising the poorest agent to 83 and 84, but their two poorest still do most of the cleaning. \alg{SRR-IA+V} is the exception. All 5 agents' returns climb together, every agent takes about a fifth of the apples and a fifth of the cleaning, and its Nash welfare, 270, equals its average return, at a collective return about 12\% below \alg{TR-EI}'s. Its high welfare comes from sharing the work, not from every agent earning equally little.

For every operator, the best self-referenced learner divides the return more evenly than the true-reward learner with the same operator, and across the whole Clean Up tuning grid, selecting on collective return picks a true-reward learner while selecting on Nash welfare picks one of ours. What spreads the return is not the self-referenced estimate alone but where it is used: the reward learners with EI and SVO use the same estimates as \alg{SRA}, yet split into eaters and cleaners just as the true-reward learners do. The other games show the same tension in their own form. In Commons Harvest the extra apples of the true-reward learner go almost entirely to one agent, leaving its Nash welfare below that of plain PPO, although welfare there needs care, since a control that ignores the others scores highest.In Escape Room every learner that escapes leaves the same agents paying for the lever in every episode. A standing ledger of self-referenced returns across episodes spreads this cost, at about a quarter of the optimum, because nothing tells the agents who should pay this time (See Supplementary Appendix F).\looseness=-1
~\vspace{-8pt}
\subsubsection{What if Agents See Only Part of the Map?}
\label{sec:results-po}
With a partial view the assessment itself becomes harder: the transition that agent $i$ receives from $j$ shows only $j$'s neighbourhood, a part of the map $i$ may never have seen from there. Figure~\ref{fig:partial-obs} shows Clean Up and Commons Harvest with this window.\looseness=-1

SRA keeps its cooperation. \alg{SRA-EI} holds its full-view production in Clean Up and still produces several times as much as plain PPO in Commons Harvest. Memory helps in Clean Up, where its recurrent agents produce more and divide the return best of all learners, while the true-reward learner again leaves its poorest agent with nothing.\looseness=-1
Inequity aversion in the reward collapses, and the reason is instructive. The estimates are not to blame; they are as accurate as with the full view. The look-ahead is. Removing it partly restores cooperation, and adding it to agents given the true rewards destroys theirs too. The look-ahead reads the critic's value off a window that shows a different part of the map, and an agent averse to being ahead reacts to a gap that is not there. The component that rescued inequity aversion under the full view is the one that breaks it when agents see less, while in the policy update an error adds noise to the credit. Appendix G gives the details.\looseness=-1
\vspace{-5pt}
\subsubsection{What if Agents Value Things Differently?}
\label{sec:results-hetero}
We now violate the condition under which self-reference can recover the other's reward. Two of five Clean Up agents value an apple at 2 rather than 1, so a self-referenced agent cannot recover the other's true reward from its transition: it assigns the transition the value that it would itself receive. Self-reference remains useful nonetheless: for every operator, our learner divides the return more evenly than the true-reward learner and leaves its poorest agent better off, at a similar collective return. What no learner does, with true or self-referenced rewards, is divide by value. The high-value agents eat about as many apples as the others, so they end with about twice the return whoever is learning. Whether self-reference misleads agents that would otherwise act on values, as people are misled when they assume others value what they value~\cite{Ross1977,Epley2004}, remains open. Supplementary Appendix H gives the details.\looseness=-1
~\vspace{-20pt}
\section{Related Work}
\label{sec:related}
\paragraph{Social preferences without reward access.}

Social preferences such as inequity aversion~\cite{Hughes2018},
social value orientation~\cite{McKee2020}, and prosocial rewards~\cite{Peysakhovich2018}
directly use other agents' rewards or outcomes. Forward-looking social
utilities provide another way to use these outcomes: AFP combines
altruistic and fairness preferences through a CES utility applied to
expected future returns~\cite{Wei2026}. Unlike our method, it still
requires access to other agents' rewards; we instead replace the peer-reward
input with a self-referenced estimate. Related approaches reshape incentives
through reward transfer or learned incentives~\cite{Lupu2020,Yang2020,Willis2024}.
Social influence~\cite{Jaques2019} also avoids access to others' rewards by
rewarding causal influence over their actions, including in a decentralized
variant that learns a model of other agents from observed actions. Unlike our
method, however, influence is the social signal itself rather than a
representation of the other's outcome. We keep the social preferences
unchanged and replace only their input.\vspace{-5pt}

\paragraph{Modeling others from one's own experience.}
Empathy is often described in terms of taking another's perspective, by projecting oneself into their situation and using one's own representations to understand their state~\cite{Preston2002}. This intuition is closely related to Simulation Theory, which proposes that people can understand others' mental states by simulating those states using their own cognitive mechanisms~\cite{Gordon1986}. In multi-agent learning, agents have inferred others' goals with their own policy~\cite{Raileanu2018}, evaluated their own value function from another's perspective~\cite{Bussmann2019}, and used perspective-taking to decide how much reward to give away~\cite{Kong2024LASE}. We use the same idea for standard social preferences and ask where the estimate should enter learning. Inverse RL~\cite{Ng2000,Natarajan2010} and inverse planning~\cite{Baker2009} infer another's reward or goals from its behavior under an assumed decision model; self-reference instead assesses the other's situation with the agent's own reward model, without requiring the other agent to act optimally. Unlike methods that keep rewards private but exchange action suggestions~\cite{Jin2025}, our agents exchange no messages. Other approaches learn social reward models from external evaluations of collective outcomes~\cite{Caftory2026}. Our agents receive no such external evaluation and instead construct social signals solely from their own reward experience. Appendix J extends this review.~\looseness=-1
~\vspace{-20pt}
\section{Conclusion and Future Work}
\label{sec:discussion}
Social preferences do not require agents to directly observe each other's rewards. An agent can instead assess another agent's situation from its own experience, by applying its learned reward model to the other's observed transitions. When reward functions are shared, this assessment can approximate the other's realized reward; when preferences differ, it instead provides a self-referenced counterfactual assessment. In both cases, the resulting signal can sustain cooperation in sequential social dilemmas.

Several directions remain open. Agents whose rewards come from different events are beyond a model of one's own reward, and an agent could instead keep its own model as a prior and correct it from what the others do, combining self-reference with inverse reinforcement learning. SRR needs no actor-critic, so it could carry self-referenced preferences to value-based learners. Letting policies see the record of past episodes, or communicate, could turn sharing into deliberate turn-taking~\cite{Axelrod1981}. A look-ahead that judges another agent's situation from memory rather than from a single window could make SRR robust to partial views. Finally, the setting fits mixed groups of people and machines, where an agent must cooperate with others whose rewards it can never observe.

\section{Acknowledgment}
We would like to thank Rafael M. Cheang for the discussions that led to the initial conception of the ideas presented in this paper, as well as for his valuable insights and help with their formulation. We are also grateful to Prof. James R. Wright, whose course, Modelling Human Strategic Behaviour, provided the setting for the first draft of this work. His feedback during the course, helped shape, expand, and formalize the ideas presented here. Finally, we thank Prof. Marlos C. Machado for his continued support and valuable feedback, as well as for providing computational resources that enabled us to further develop and finalize this project.






\bibliographystyle{ACM-Reference-Format} 
\bibliography{references}
\clearpage
\appendix

These appendices give the details and the evidence behind the paper.
Appendices~\ref{app:method-details} and~\ref{app:exp-details} describe the
implementation and the experimental protocol.
Appendices~\ref{app:res-cooperate} to~\ref{app:hetero} extend the
subsections of results, one each and in the same order.
Appendix~\ref{app:limitations} collects the limitations of the study, and
Appendix~\ref{app:related} reviews the related work in full. Figures,
tables, and equations numbered with an S belong to the appendices.

\FloatBarrier
\section{Method Details}
\label{app:method-details}

This appendix elaborates on the methods section. Specifically we discuss why we can use the social terms within the coefficient of the policy update, the order of one training iteration, how the reward model is trained, how the scores are scaled, and how episode boundaries are handled.
\subsection{Policy Gradients and the Baseline}
\label{app:pg}

The background section uses two facts about policy gradients: the
advantage measures how much better an action went than expected, and a term
that does not depend on the action leaves the expected gradient unchanged.
Both follow from the policy gradient theorem~\cite{Sutton2000,Williams1992},
\begin{equation}
  \nabla J(\theta_i) = \mathbb{E}\Bigl[\sum_t \nabla \log \pi_i(a_i^t \mid o_i^t)\,
  \bigl(G_i^t - b(o_i^t)\bigr)\Bigr],
  \label{eq:pg}
\end{equation}
where $G_i^t = \sum_{k \ge 0} \gamma^k r_i^{t+k}$ is the return that followed
action $a_i^t$ and $b$ is a \emph{baseline}, any function that does not
depend on the action. The baseline does not change the gradient in
expectation, because
\begin{equation}
  \sum_a b(o)\, \nabla \pi_i(a \mid o) = b(o)\, \nabla \sum_a \pi_i(a \mid o)
  = b(o)\, \nabla 1 = 0 ,
  \label{eq:baseline}
\end{equation}
but it reduces the variance of the estimate~\cite{Sutton2018}. Actor-critic
methods use a learned critic $V_i(o) \approx \mathbb{E}[G_i^t \mid o_i^t = o]$
as the baseline and estimate the advantage $G_i^t - V_i(o_i^t)$ by
bootstrapping. With the temporal-difference error, the generalized
advantage estimate~\cite{Schulman2016} is
\begin{align}
  \delta_i^t &= r_i^t + \gamma V_i(o_i^{t+1}) - V_i(o_i^t), \nonumber\\
  A_i^t &= \sum_{k \ge 0} (\gamma\lambda)^k \delta_i^{t+k},
  \label{eq:gae}
\end{align}
where $\lambda \in [0, 1]$ trades bias for variance; with $\lambda = 1$,
$A_i^t = G_i^t - V_i(o_i^t)$. Eq.~\eqref{eq:baseline} is why the
coefficient of the policy update is a separate place for a social term: a
term added to $A_i^t$ changes the expected update only through its
dependence on the action. It is also why \alg{SRA} is not simply a
reweighted version of \alg{SRR}. The cross scores it adds depend on another
agent's action, not on $i$'s, so they do not vanish in expectation, but they
also do not correspond to any fixed objective that $i$ maximizes.

\subsection{One Training Iteration}
\label{app:ssa-details}

Table~\ref{tab:rewards} summarizes what each learner optimizes. The four
rows differ in two columns only, the reward that PPO and the critic see and
the coefficient that weights the actor's update, which is the distinction
the paper turns on: \alg{SRR} changes the first and \alg{SRA} the second.

Algorithm~\ref{alg:method} gives one training iteration of agent $i$. The
order matters in three places. The rollout is collected with the networks
and reward models frozen, so every self-referenced estimate within an iteration comes from the same model. The \alg{SRR} term and the standing gate are applied in
the order of the transitions, with the standing from before the current
episode, so an agent is never penalized for a lead it acquires in the
episode being scored. And all networks are updated only after the social
coefficients of the rollout are fixed, so a reward model is used from the
rollout after the one it was fitted on. Evaluating $N$ reward models and
critics on $N$ streams costs $\mathcal{O}(N^2)$ forward passes per step,
which is the main cost of the method over independent PPO.

\begin{table}[tb]
  \centering
  \caption{The learning signal of each learner. $V_i^{\rm obj}$ is trained
  on the return of the reward in the second column.}
  \label{tab:rewards}
  \centering
  \small
  \setlength{\tabcolsep}{3pt}
  \begin{tabular}{@{}llll@{}}
    \toprule
    Learner & PPO reward & actor coefficient & $f_i$ target \\
    \midrule
    Plain PPO & $r_i$ & own GAE & -- \\
    \alg{TR} & $r_i+F_i(x^{\rm true})$ & own GAE & -- \\
    \alg{SRR} & $r_i+F_i(x^{\rm pred})$ & own GAE & $r_i$ \\
    \alg{SRA} & $r_i$ & $\widetilde\Psi_i$ & $r_i$ \\
    + Standing & above $-\kappa\,\mathrm{ahead}_i[r_i]_+$ & unchanged & unchanged \\
    \bottomrule
  \end{tabular}
\end{table}

\begin{algorithm}[tb]
  \caption{One training iteration for agent $i$.}
  \label{alg:method}
  \small
  \begin{algorithmic}[1]
    \State Collect all transition streams; reveal only $r_i$ to learner $i$
    \State Evaluate $\hat r_{ij}^t\gets f_i(o_j^t,a_j^t,o_j^{t+1})$; set $\hat r_{ii}^t\gets r_i^t$
    \If{reward-model warm-up is complete}
      \State For \alg{SRR}, compute $x_{ij}$ and $u_i\gets r_i+F_i(x_{i\cdot})$
      \State Apply standing adjustment using the pre-update ledger; then update $G_i$ with completed episodes
    \Else
      \State $u_i\gets r_i$
    \EndIf
    \State Compute own GAE from $u_i$; for \alg{SRA} compute $\hat A_{ij}$ and $\widetilde\Psi_i$
    \State Optimize actor and objective critic; regress $f_i$ on own current and replay transitions
  \end{algorithmic}
\end{algorithm}

\subsection{Reward-Model Training}
\label{app:rm}

Each reward model is trained on the current minibatch and, once its replay
holds any transitions, on an independently sampled replay minibatch,
\begin{equation}
 \mathcal{L}^{\mathrm{RM}}_i
 =\frac{1}{2}\mathbb{E}_{\mathcal{B}_i}(f_i-r_i)^2
 +\mathbf{1}[|\mathcal{R}_i|>0]\,
  \frac{1}{2}\mathbb{E}_{\mathcal{R}_i}(f_i-r_i)^2.
 \label{eq:app-rm-loss}
\end{equation}
The replay belongs to agent $i$ alone, holds only its own transitions with
a nonzero reward, up to 2{,}048 of them, and is sampled with replacement.
Its purpose is coverage rather than efficiency: an agent that settles into
cleaning the river stops eating apples, and without the replay its
minibatches would soon contain no example of an apple reward, so it would
forget how to recognize one in another agent's transition. The estimates that enter the social terms are computed before the update and carry no gradient into the policy, so the reward model is fitted only to the agent's own rewards and never to make the social term convenient. The social term stays
off for the first $W$ environment steps, 20k in the three environments, while the model is fitted.

With value clipping, the critic loss is
\begin{equation}
 \begin{split}
 \mathcal{L}_{V_i}=\frac{1}{2}\mathbb{E}\max\{&
 (V_i-R_i)^2,\\
 &[V_i^{\mathrm{old}}+
 \operatorname{clip}(V_i-V_i^{\mathrm{old}},
                     -\epsilon_c,\epsilon_c)-R_i]^2\},
 \end{split}
 \label{eq:app-value-loss}
\end{equation}
and each agent minimizes
\begin{equation}
 \begin{split}
 \mathcal{L}_i={}&\mathcal{L}^{\mathrm{actor}}_i
 +c_V\mathcal{L}_{V_i}
 -c_H\mathbb{E}[\mathcal{H}(\pi_i)]\\
 &+\mathbf{1}_{\mathrm{raw}}c_V\mathcal{L}_{V_i^{\mathrm{raw}}}
 +\mathbf{1}_{\mathrm{SR}}c_f\mathcal{L}^{\mathrm{RM}}_i,
 \end{split}
 \label{eq:app-joint-loss}
\end{equation}
where $\mathcal{L}^{\mathrm{actor}}_i$ is PPO's clipped objective with the
coefficient of Table~\ref{tab:rewards} in place of the advantage, the raw
critic $V_i^{\mathrm{raw}}$ appears only in the control of
Appendix~\ref{app:standing-cost}, and the reward-model term only for
self-referenced learners. The actor and the critic of an agent share its
observation encoder, no parameters are shared between agents, and
gradients are clipped for each agent separately.

\subsection{Scaling the Scores}
\label{app:scaling}

In \alg{SRA}, the agent's own score and each cross score are standardized
over the time steps and parallel environments of a rollout before they are
combined,
\begin{equation}
 \mathcal{S}_{\mathrm{roll}}(x)
 =\frac{x-\mu_{t,e}(x)}
 {\max(\sigma_{t,e}(x),10^{-6})},
 \label{eq:app-rollout-scale}
\end{equation}
so a constant score becomes zero, and PPO normalizes the combined
coefficient once more within each minibatch. Standardizing first is what
makes a single weight $\alpha$ meaningful across games: the cross scores of
an agent in Commons Harvest, where a beam costs 50, would otherwise dwarf
those in Escape Room, where a lever costs 1. The price is that $\alpha$ and
$\beta$ in \alg{SRA} weight standardized scores rather than rewards, so
their values are not comparable with those of \alg{SRR} or \alg{TR}, and a
larger weight in one is not a stronger preference in the other. The social
terms average over the other agents, dividing by $N-1$; a variant that takes
the largest gap instead of the mean is not used in any reported result.

\subsection{Episode Boundaries}
\label{app:boundaries}

An episode ends either because the agents finish or because of a time limit, and the two need different treatment. At a true end there is nothing to bootstrap, so the cross score defined in the main paper uses no next value and its recursion stops. At a time limit the episode could have continued, so the score keeps the value of the last observation, but the recursion still stops, so that one episode is never linked to the next. The look-ahead of \texttt{SRR+V} is dropped only at true ends, where there is no future to anticipate. For the agent's own advantage we add the value of the last observation to the final reward at a time limit, which gives the same result as bootstrapping it. Getting this wrong matters most in Escape Room, where episodes are five steps long, so a score that leaked across boundaries would mix the outcome of one attempt into the next.

\section{Experimental Details}
\label{app:exp-details}

This appendix describes what the agents observe and can do, how they are built and trained, how every reported setting was chosen, and how the partial-view experiments differ from the main ones.
\subsection{Observations and Actions}
\label{app:observations}

\begin{figure*}[tp]
  \centering
  \includegraphics[width=\textwidth]{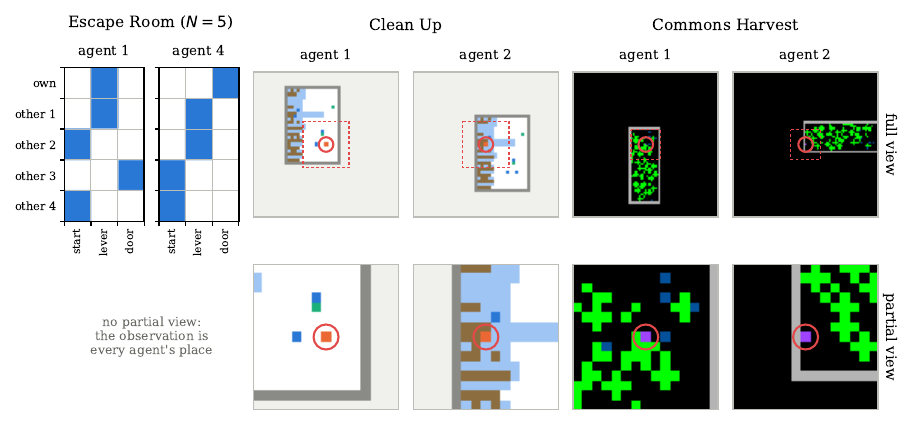}
  \caption{One state of each game as two agents observe it, after a few
  random steps, with the full view (top) and the partial view (bottom).
  Escape Room ($N = 5$): the one-hot place of every agent, the observer's
  own first and the others in a fixed order; agent 1 is at the lever and
  agent 4 at the door. The observation is not spatial, so it has no partial
  version. Clean Up: the $47 \times 47$ window centred on the observer (red
  circle), large enough to contain the whole $25 \times 18$ map from any
  position, shown with one colour per plane (orange self, blue others, green
  apples, brown waste, light blue river, grey walls; beams yellow). Commons
  Harvest: the $73 \times 73$ RGB window centred on the observer and turned
  to the direction it faces, so the two agents see the map at different
  angles; apples are green, the observer and the others have one fixed
  colour each. The partial view is the central $15 \times 15$ part of the
  same window (dashed box, top), used in the partial-view experiments of Appendix~\ref{app:po}.
  Everything outside the map is empty.}
  \label{fig:app-observations}
\end{figure*}

Table~\ref{tab:envs} lists what each agent observes and can do, and
Figure~\ref{fig:app-observations} shows one state of each game as two
different agents see it, side by side, with the full view in the top row
and the partial view below.

The figure makes concrete the property the method depends on: every
observation is arranged around its own agent, so the same reward model and
critic can be applied to any agent's observation. In Escape Room the
observation is a stack of one-hot rows, the observer's own place first and
the others below in a fixed order; agents 1 and 4 therefore see the same
state as two different matrices. In Clean Up the window is centred on the
observer, drawn here with one colour per plane, and in Commons Harvest it is
centred and also turned to the direction the agent faces, so the two
columns show the same map at different angles.

The bottom row shows what the partial view removes. With the full view the
window is large enough to contain the whole map from any position, so agent
$i$ sees in $o_j$ everything agent $j$ sees, and assessing $j$'s transition is a matter of recognizing the event. With the $15 \times 15$ window, much of
$o_j$ lies outside $i$'s own window: in the example, agent 1's Clean Up
window contains the river bank and two other agents, while agent 2's
contains mostly river and waste. Agent $i$ must then judge $j$'s situation
from a view it has never had itself, which is the difficulty that Appendix~\ref{app:po} examines. Escape Room has no partial version, because
its observation is already the list of places rather than a map.

\begin{table}[tb]
  \centering
  \caption{Observation and actions of one agent. Every observation is
  arranged relative to its agent. The full view covers the whole state and
  is used in the main experiments; the partial view keeps only the
  $15 \times 15$ neighbourhood of the agent. Figure~\ref{fig:app-observations} shows
  examples of both.}
  \label{tab:envs}
  \centering
  \small
  \setlength{\tabcolsep}{3pt}
  \begin{tabular}{@{}lp{0.46\columnwidth}p{0.24\columnwidth}@{}}
    \toprule
    Game & Observation & Actions \\
    \midrule
    Escape Room & one-hot place (start, lever, door) of every agent, its own first; $3N$ values; no partial view & 3: stay, lever, door \\
    Clean Up & window centred on the agent, 7 binary planes: self, others, apple, waste, river, wall, beam; $47 \times 47$ full, $15 \times 15$ partial & 6: four moves, stay, clean \\
    Commons Harvest & RGB window centred on the agent and turned to its facing direction, self and others each in one fixed colour; $73 \times 73$ full, $15 \times 15$ partial & 8: four moves, stay, two turns, fire \\
    \bottomrule
  \end{tabular}
\end{table}

\subsection{Networks and PPO Settings}
\label{app:settings}

Table~\ref{tab:app-settings} lists the networks and PPO settings of every
game, and Table~\ref{tab:app-er-ppo} the PPO setting of each Escape Room
learner with five agents. All learners use $\gamma = 0.99$, GAE
$\lambda = 0.95$, clip $0.2$, four epochs of four minibatches per update,
value coefficient $0.5$, gradient-norm clip $0.5$, and a linearly annealed
learning rate.

The reward model is the one component we had to adapt per game. In Escape
Room and Clean Up it is a separate network on the transition, small enough
to fit quickly. In Commons Harvest, where the observation is a
$73 \times 73$ image, a separate convolutional network was slow to fit from
the few rewarded transitions available, so the model is instead a small head
on the policy encoder's features of $o$ and $o'$. The features are detached,
so fitting the reward model does not change the encoder, and the policy is
unaffected by the reward model except through the social term.

\begin{table*}[tp]
  \centering
  \caption{Networks and PPO settings. The actor and the critic are linear
  heads on one shared network. Escape Room tunes the PPO setting per
  learner (Table~\ref{tab:app-er-ppo}); Clean Up and Commons Harvest use
  one setting for all learners, chosen in the PPO stage of the selection
  protocol.}
  \label{tab:app-settings}
  \centering
  \small
  \begin{tabular}{@{}lp{0.26\textwidth}p{0.26\textwidth}p{0.26\textwidth}@{}}
    \toprule
    & Escape Room & Clean Up & Commons Harvest \\
    \midrule
    Shared network & MLP, two layers of 64 tanh units & $3 \times 3$ convolutions with 16 and 32 channels (second with stride 2), 128 ReLU units & CleanRL Atari network: convolutions $8 \times 8$/4 (32), $4 \times 4$/2 (64), $3 \times 3$/1 (64), 512 ReLU units \\
    Reward model $f_i$ & MLP, two layers of 64 tanh units, on $(o, a, o')$ & two $3 \times 3$ convolutions (16 channels) on the stacked $(o, o')$, max- and mean-pooled, then 64 units with $a$ & one layer of 64 tanh units on the shared network's features of $o$ and $o'$ (not trained through) and $a$ \\
    Steps; parallel envs & 500k; 16 & 5M; 8 & 5M; 8 \\
    Rollout per env & 16, 32, or 128 (tuned) & 256 & 128 \\
    Learning rate & $2.5 \times 10^{-4}$ or $10^{-3}$ (tuned) & $10^{-3}$ & $10^{-3}$ \\
    Entropy coefficient & 0.01 or 0.05 (tuned) & 0.01 ($N = 5$); 0.05 ($N = 3$)$^\ast$ & 0.01 \\
    PPO stage & joint with the operator weights & learning rate $\{2.5 \times 10^{-4}, 10^{-3}\}$ $\times$ entropy $\{0.01, 0.05\}$ $\times$ rollout $\{128, 256, 512\}$ on plain PPO and six social configurations, three seeds & learning rate $\{2.5 \times 10^{-4}, 10^{-3}\}$ $\times$ entropy $\{0.003, 0.01, 0.03\}$ on plain PPO, two seeds \\
    Warm-up $W$ & 20k steps & 20k steps & 20k steps \\
    Tuning; held-out seeds & 1--3; 11--18 & 1--3; 7--14 & 1--3; 4--11 \\
    \bottomrule
  \end{tabular}
  \par\smallskip
  \footnotesize $^\ast$At $N = 3$ plain PPO uses its own tuned setting: learning rate $2.5 \times 10^{-4}$, entropy 0.01, rollout 128.
\end{table*}

\begin{table}[tb]
  \centering
  \caption{Escape Room, $N = 5$: PPO setting of each learner at its
  collective-return pick (learning rate, entropy coefficient, rollout
  length per environment).}
  \label{tab:app-er-ppo}
  \centering
  \small
  \begin{tabular}{@{}llr@{}}
    \toprule
    Learner (weights) & LR, entropy & Rollout \\
    \midrule
    Plain PPO & $10^{-3}$, 0.01 & 32 \\
    \alg{TR-EI} ($\alpha = 1$) & $10^{-3}$, 0.01 & 128 \\
    \alg{TR-IA} ($\alpha/\beta = 0/1$) & $10^{-3}$, 0.01 & 32 \\
    \alg{TR-SVO} ($\alpha = 1$, $\phi = \pi/4$) & $10^{-3}$, 0.01 & 128 \\
    \alg{SRA-EI} ($\alpha = 3$) & $2.5 \times 10^{-4}$, 0.05 & 128 \\
    \alg{SRA-IA} ($\alpha/\beta = 0/0.5$) & $10^{-3}$, 0.05 & 128 \\
    \alg{SRA-SVO} ($\alpha = 5$, $\phi = \pi/3$) & $2.5 \times 10^{-4}$, 0.01 & 128 \\
    \alg{SRR-EI} ($\alpha = 0.1$) & $10^{-3}$, 0.05 & 128 \\
    \alg{SRR-IA} ($\alpha/\beta = 0/1$) & $2.5 \times 10^{-4}$, 0.05 & 128 \\
    \alg{SRR-IA+V} ($\alpha/\beta = 0/1$) & $2.5 \times 10^{-4}$, 0.01 & 128 \\
    \alg{SRR-IA} + Standing ($\kappa = 1$) & $2.5 \times 10^{-4}$, 0.05 & 128 \\
    \bottomrule
  \end{tabular}
\end{table}

\subsection{Hyperparameter Selection}
\label{app:selection}

Every number in the paper comes from settings chosen before the held-out
runs. In Clean Up and Commons Harvest we first choose one PPO setting and
keep it for every learner, so that no learner is favoured by a better base
configuration; in Escape Room, where runs are cheap, we tune the PPO setting
together with each learner's social weights. We then score every candidate
by the mean minus the sample standard deviation of the metric across the
tuning seeds, once with collective return and once with Nash welfare, on
seeds 1--3 in every game.
Subtracting one standard deviation favours settings that work on every
tuning seed rather than settings with one good seed; with two or three seeds
it is a rule of thumb, not a confidence bound. Each chosen setting is then
trained from scratch on eight held-out seeds, 11--18 in Escape Room, 7--14
in Clean Up, and 4--11 in Commons Harvest, and nothing from these runs feeds
back into the choice. When both metrics choose the same setting, it is run
once.

\begin{figure}[tb]
  \centering
  \includegraphics[width=\columnwidth]{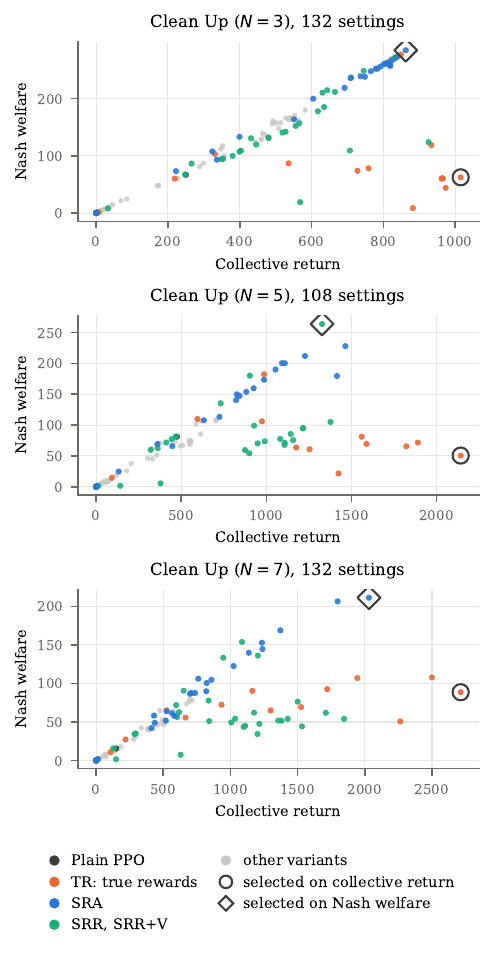}
  \caption{Clean Up tuning landscape. Each point is one setting evaluated
  on the three tuning seeds, placed by its mean collective return and mean
  Nash welfare and coloured by learner family; grey points are variants not analysed in the paper. The circle and the diamond mark the settings
  that the mean-minus-standard-deviation score selects on collective return
  and on Nash welfare, among all candidates at that group size.}
  \label{fig:app-cleanup-landscape}
\end{figure}

Figure~\ref{fig:app-cleanup-landscape} shows why we select twice. Each point is one Clean Up setting, placed by its mean collective return and mean Nash welfare over the tuning seeds, with one panel per group size and one colour per learner family. For most of the grid the two metrics agree: the cloud rises along a diagonal, where a setting that produces more also divides it more evenly, simply because the alternative to producing is that nobody eats. The cloud then splits. Above about 800 apples with three agents and 1{,}000 with five, one branch keeps rising and the other turns right and down, toward settings that produce more but leave some agents with little.

The colours show who is on each branch. The branch that turns down holds the true-reward learners, in orange, and many reward-placement settings, in green; the branch that keeps rising is mostly \alg{SRA}, in blue, joined by the best \alg{SRR+V} settings. At every group size the setting selected on collective return, the circle, is a true-reward learner at the far right with a Nash welfare near the bottom of the range, and the setting selected on Nash welfare, the diamond, is a self-referenced learner at the top: \alg{SRA-EI} with three and seven agents and \alg{SRR-IA+V} with five. No setting is at the corner of both, which is why we report both selections throughout rather than choosing one.

The grids hold 132 candidates in Clean Up with three agents, 108 with five,
132 with seven, and 40 in Commons Harvest, including the self-scaling
controls; in Escape Room they cover every operator, placement, and standing
setting. Within a learner, the two metrics choose different settings for 7
of 21 learners with three agents, 5 of 21 with five, and 6 of 21 with seven,
and in Escape Room for 15 of 126 combinations of group size and learner.
Where they differ, the setting chosen on Nash welfare often has a slightly
lower mean but varies much less across seeds.

\subsection{Partial View}
\label{app:po-settings}

For the partial view, every learner is retuned and trained for 7 million
environment steps, more than the 5 million of the full view, because the
smaller window makes the task slower to learn for every learner including
plain PPO. The grid scales each learner's full-view social weights by
$\tfrac13$, 1, and 3 and crosses them with learning rates $10^{-3}$ and
$3\times10^{-4}$ and entropy coefficients 0.01 and 0.05, on tuning seeds 1
and 2; the selected setting is rerun on the same held-out seeds as with the
full view. Scaling the full-view weights rather than searching from scratch
keeps the grid small, and it assumes that the right weight moves with the
view but not far, which the selected settings bear out.

The recurrent agents carry a GRU~\cite{Cho2014} in the policy and in the
critic over the agent's observation, its previous action, and its previous
reward, with layer normalization, a skip connection around the recurrent
layer, and truncated backpropagation over 64 steps. The reward model stays
feed-forward and reads only the transition, so memory can help the agent
act and estimate values but is not available to the self-referenced estimate itself.

\FloatBarrier
\section{Can Agents Cooperate without Observing Others' Rewards?}
\label{app:res-cooperate}

Can social preferences be grounded in self-referenced estimates of others' outcomes rather than in their reward signals? This appendix gathers the evidence. It places our learners among all the learners we ran, gives the full results, adds the two measures that collective return leaves out, and asks how far the Escape Room result extends. Two conclusions emerge. Self-referenced estimates are enough to start and sustain cooperation in all three games. Where they fall short of the true rewards, in Commons Harvest, the extra production of the true-reward learners goes to a few agents.

\subsection{Where Our Learners Stand}
\label{app:overview}

\begin{figure}[tb]
  \centering
  \includegraphics[width=\columnwidth]{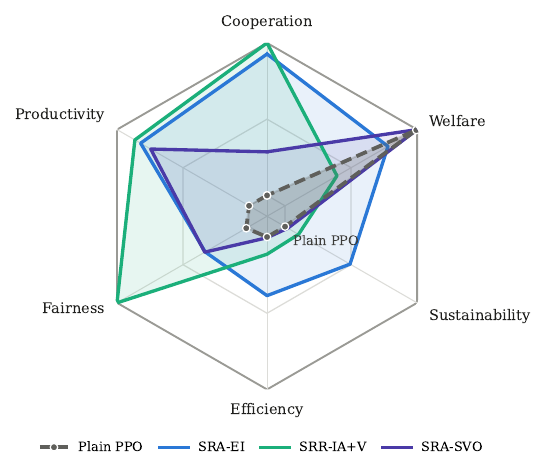}
  \caption{Our three self-referenced learners, which never observe another
  agent's reward, against plain PPO on what each game tests: cooperation in
  Escape Room, productivity and fairness in Clean Up, and efficiency,
  sustainability, and welfare in Commons Harvest. Each axis spans the means
  of all learners in that game, true-reward ones included; in Commons
  Harvest the true-reward agents reach further on efficiency and
  sustainability. Eight held-out seeds.}
  \Description{Radar plot comparing plain PPO with three self-referenced learners on six properties across the three games.}
  \label{fig:intro}
\end{figure}

Figure~\ref{fig:intro} places our three self-referenced learners among all the learners we ran, on what each game tests:
cooperation in Escape Room, productivity and fairness in Clean Up, and
efficiency, sustainability, and welfare in Commons Harvest. Each axis spans
the means of all learners in that game, true-reward learners included, so
reaching the edge means doing as well as the best learner we ran, not merely
better than plain PPO.

Plain
PPO is the small region in the middle; it reaches the welfare axis only
because agents that each earn almost nothing are equal. Our learners extend
it in different directions, \alg{SRA-EI} towards production and
\alg{SRR-IA+V} towards fairness, and between them they reach the edge on
most axes. The two axes where the true-reward learners still lead,
efficiency and sustainability in Commons Harvest, are the shortfall that the next figures trace to how the return is divided.

\subsection{The Full Results}
\label{app:tables}

Table~\ref{tab:selection} gives the final collective return, Nash welfare,
and lowest agent return of every learner in every game, at both of its
selections. Table~\ref{tab:harvest} adds sustainability for Commons Harvest,
where a high collective return can also come from stripping the map early,
and Table~\ref{tab:app-headline-n5} adds a behavioural measure for Escape
Room and Clean Up with five agents: how spread the exit rates are, and how
much of the cleaning the two poorest agents do.

Two entries in Table~\ref{tab:harvest} change how the Commons Harvest
results should be read. \alg{TR-IA} ends with a collective return of
$-1{,}935$, which only the beam can produce, so its agents are not merely
failing to harvest but firing at one another. And the self-scaling control,
which rescales the agent's own reward and uses no information about the
others, has the highest Nash welfare and the highest lowest agent return of
any learner in the game. A high welfare in this game can therefore be
reached without any preference over the others, which is why we report the
control separately and do not read welfare there as evidence of cooperation.

\begin{table*}[t]
  \caption{Every learner at the hyperparameter setting selected on tuning seeds by collective return and by Nash welfare (``same setting'' when both objectives selected it). $C$, $\NW$, and $L$ are computed from each seed's final-window per-agent means. Entries are held-out-seed means $\pm$ 95\% Student-$t$ intervals; they describe the frozen selections and are not used to reselect them.}
  \label{tab:selection}
  \centering
  \footnotesize
  \setlength{\tabcolsep}{3pt}
  \begin{tabular}{@{}lrrrrrr@{}}
    \toprule
    & \multicolumn{3}{c}{selected on collective return} & \multicolumn{3}{c}{selected on Nash welfare} \\
    \cmidrule(lr){2-4}\cmidrule(l){5-7}
    Learner & $C$ & $\NW$ & $L$ & $C$ & $\NW$ & $L$ \\
    \midrule
    \multicolumn{7}{@{}l}{\emph{Escape Room ($N=5$)}} \\
    Plain PPO & $-0.3 \pm 0.4$ & $4.95 \pm 0.08$ & $-0.25 \pm 0.39$ & \multicolumn{3}{c}{same setting} \\
    \addlinespace[2pt]
    \alg{TR-EI} & $\mathbf{17.0 \pm 0.0}$ & $\mathbf{6.96 \pm 0.00}$ & $-1.00 \pm 0.00$ & \multicolumn{3}{c}{same setting} \\
    \alg{SRA-EI} & $15.7 \pm 3.0$ & $6.78 \pm 0.42$ & $-1.00 \pm 0.00$ & \multicolumn{3}{c}{same setting} \\
    \alg{SRR-EI} & $0.0 \pm 0.0$ & $5.00 \pm 0.00$ & $0.00 \pm 0.00$ & \multicolumn{3}{c}{same setting} \\
    \addlinespace[2pt]
    \alg{TR-IA} & $\mathbf{17.0 \pm 0.0}$ & $\mathbf{6.96 \pm 0.00}$ & $-1.00 \pm 0.00$ & $\mathbf{17.0 \pm 0.0}$ & $\mathbf{6.96 \pm 0.00}$ & $-1.00 \pm 0.00$ \\
    \alg{SRA-IA} & $0.0 \pm 0.0$ & $5.00 \pm 0.00$ & $0.00 \pm 0.00$ & \multicolumn{3}{c}{same setting} \\
    \alg{SRR-IA+V} & $\mathbf{17.0 \pm 0.0}$ & $\mathbf{6.96 \pm 0.00}$ & $-1.00 \pm 0.00$ & \multicolumn{3}{c}{same setting} \\
    \addlinespace[2pt]
    \alg{TR-SVO} & $14.9 \pm 5.0$ & $6.72 \pm 0.58$ & $-0.88 \pm 0.30$ & \multicolumn{3}{c}{same setting} \\
    \alg{SRA-SVO} & $4.7 \pm 8.2$ & $5.41 \pm 1.14$ & $-1.00 \pm 0.00$ & \multicolumn{3}{c}{same setting} \\
    \alg{SRR-SVO} & $0.0 \pm 0.0$ & $5.00 \pm 0.00$ & $0.00 \pm 0.00$ & \multicolumn{3}{c}{same setting} \\
    \midrule
    \multicolumn{7}{@{}l}{\emph{Clean Up ($N=5$)}} \\
    Plain PPO & $248 \pm 198$ & $37.0 \pm 29.8$ & $5.4 \pm 4.8$ & \multicolumn{3}{c}{same setting} \\
    \addlinespace[2pt]
    \alg{TR-EI} & $1{,}530 \pm 289$ & $70.4 \pm 33.6$ & $0.0 \pm 0.0$ & $1{,}159 \pm 444$ & $86.7 \pm 14.6$ & $3.3 \pm 5.2$ \\
    \alg{SRA-EI} & $1{,}286 \pm 141$ & $214.4 \pm 15.3$ & $83.1 \pm 16.6$ & \multicolumn{3}{c}{same setting} \\
    \alg{SRR-EI} & $1{,}512 \pm 140$ & $84.5 \pm 29.8$ & $0.0 \pm 0.0$ & \multicolumn{3}{c}{same setting} \\
    \addlinespace[2pt]
    \alg{TR-IA} & $964 \pm 107$ & $178.1 \pm 20.6$ & $95.9 \pm 18.0$ & \multicolumn{3}{c}{same setting} \\
    \alg{SRA-IA} & $590 \pm 188$ & $101.6 \pm 38.5$ & $39.5 \pm 32.8$ & \multicolumn{3}{c}{same setting} \\
    \alg{SRR-IA+V} & $1{,}353 \pm 93$ & $\mathbf{269.9 \pm 18.8}$ & $251.5 \pm 23.4$ & \multicolumn{3}{c}{same setting} \\
    \addlinespace[2pt]
    \alg{TR-SVO} & $1{,}405 \pm 316$ & $64.6 \pm 37.8$ & $5.3 \pm 12.5$ & $\mathbf{1{,}710 \pm 499}$ & $60.6 \pm 26.7$ & $0.0 \pm 0.0$ \\
    \alg{SRA-SVO} & $1{,}205 \pm 39$ & $209.6 \pm 9.2$ & $83.7 \pm 13.0$ & \multicolumn{3}{c}{same setting} \\
    \alg{SRR-SVO} & $1{,}357 \pm 192$ & $71.7 \pm 32.0$ & $0.0 \pm 0.0$ & $1{,}309 \pm 293$ & $105.3 \pm 40.1$ & $15.7 \pm 30.8$ \\
    \midrule
    \multicolumn{7}{@{}l}{\emph{Commons Harvest ($N=7$)}} \\
    Plain PPO & $211 \pm 2$ & $30.1 \pm 0.2$ & $27.9 \pm 0.7$ & \multicolumn{3}{c}{same setting} \\
    \addlinespace[2pt]
    \alg{TR-EI} & $657 \pm 86$ & $5.6 \pm 1.6$ & $0.6 \pm 0.7$ & \multicolumn{3}{c}{same setting} \\
    \alg{SRA-EI} & $416 \pm 86$ & $\mathbf{35.9 \pm 3.9}$ & $8.5 \pm 4.2$ & \multicolumn{3}{c}{same setting} \\
    \alg{SRR-EI} & $-1{,}604 \pm 4{,}721$ & $15.0 \pm 7.1$ & $-522 \pm 1{,}235$ & $214 \pm 2$ & $30.5 \pm 0.2$ & $27.6 \pm 1.0$ \\
    \addlinespace[2pt]
    \alg{TR-IA} & $-1{,}935 \pm 4{,}576$ & $0.2 \pm 0.1$ & $-771 \pm 1{,}821$ & $-1{,}678 \pm 3{,}990$ & $1.0 \pm 0.6$ & $-1{,}037 \pm 2{,}454$ \\
    \alg{SRA-IA} & $212 \pm 3$ & $30.2 \pm 0.4$ & $26.7 \pm 1.7$ & \multicolumn{3}{c}{same setting} \\
    \alg{SRR-IA+V} & $271 \pm 66$ & $15.2 \pm 5.3$ & $0.3 \pm 0.7$ & $209 \pm 1$ & $29.8 \pm 0.2$ & $27.0 \pm 1.2$ \\
    \addlinespace[2pt]
    \alg{TR-SVO} & $\mathbf{744 \pm 128}$ & $26.5 \pm 12.7$ & $3.2 \pm 7.6$ & \multicolumn{3}{c}{same setting} \\
    \alg{SRA-SVO} & $213 \pm 6$ & $30.1 \pm 1.3$ & $24.4 \pm 4.6$ & \multicolumn{3}{c}{same setting} \\
    \alg{SRR-SVO} & $280 \pm 35$ & $18.7 \pm 7.1$ & $3.5 \pm 9.0$ & $214 \pm 4$ & $30.4 \pm 0.6$ & $26.5 \pm 1.7$ \\
    \bottomrule
  \end{tabular}
\end{table*}

\begin{table}[tb]
  \centering
  \caption{Commons Harvest, seven agents, eight seeds held out from tuning
  (4--11). Values are means $\pm$ 95\% Student-$t$ interval half-widths
  where shown; bold marks the best per column. In parentheses, the metric that selected the setting (C: collective
  return; NW: Nash welfare; both: the same setting).
  Sust.: mean step at which apples are eaten. $^\ast$Means of \alg{SRA-IA} and
  \alg{SRR-IA} (both) and of \alg{SRR-EI}, \alg{SRR-SVO}, and \alg{SRR-IA+V} (NW); half-widths $\le 4.2$ ($C$), $\le 0.6$ ($\NW$), and $\le 1.7$ ($L$). $L$: lowest agent return. Self-scaling control: SVO with $\phi = 0$, which only rescales the agent's own reward.}
  \label{tab:harvest}
  \centering
  \footnotesize
  \setlength{\tabcolsep}{3pt}
    \begin{tabular}{@{}lrrrr@{}}
    \toprule
    Method (selection) & $C$ & $\NW$ & $L$ & Sust. \\
    \midrule
    Plain PPO & $211 \pm 2$ & $30.1 \pm 0.2$ & $27.9 \pm 0.7$ & $46 \pm 1$ \\
    \alg{TR-EI} (both) & $657 \pm 86$ & $5.6 \pm 1.6$ & $0.6 \pm 0.7$ & $195 \pm 55$ \\
    \alg{TR-SVO} (both) & $\mathbf{744 \pm 128}$ & $26.5 \pm 12.7$ & $3.2 \pm 7.6$ & $\mathbf{349 \pm 56}$ \\
    \alg{TR-IA} (C) & $-1{,}935$ & $0.2$ & $-771$ & $88$ \\
    \alg{TR-IA} (NW) & $-1{,}678$ & $1.0$ & $-1{,}037$ & $28$ \\
    \alg{SRA-EI} (both) & $416 \pm 86$ & $35.9 \pm 3.9$ & $8.5 \pm 4.2$ & $195 \pm 44$ \\
    \alg{SRA-SVO} (both) & $213 \pm 6$ & $30.1 \pm 1.3$ & $24.4 \pm 4.6$ & $52 \pm 5$ \\
    \alg{SRR-SVO} (C) & $280 \pm 35$ & $18.7 \pm 7.1$ & $3.5 \pm 9.0$ & $98 \pm 28$ \\
    \alg{SRR-IA+V} (C) & $271 \pm 66$ & $15.2 \pm 5.3$ & $0.3 \pm 0.7$ & $77 \pm 31$ \\
    Five NW selections$^\ast$ & 209--214 & 29.8--30.5 & 26.5--28.2 & 45--49 \\
    Self-scaling control (both) & $302 \pm 64$ & $\mathbf{42.0 \pm 6.9}$ & $\mathbf{31.2 \pm 2.4}$ & $108 \pm 47$ \\
    \bottomrule
  \end{tabular}
\end{table}

\begin{table*}[tp]
  \centering
  \caption{Escape Room and Clean Up with five agents, eight held-out seeds. $C$ is
  collective return, $\NW$ Nash welfare ($c = 5$ in Escape Room, $0$ in
  Clean Up), $L$ the lowest agent return. In parentheses, the metric that selected a setting when the two differ. The last column describes behaviour: the standard deviation of the exit rates in Escape Room, about 0.5 when the same agents always exit and 0 when all exit equally often, and in Clean Up the share of the cleaning done by the two lowest earners, 0.40 for an equal split.}
  \label{tab:app-headline-n5}
  \centering
  \small
  \begin{tabular}{@{}llrrrl@{}}
    \toprule
    Env. & Method (selection) & $C$ & $\NW$ & $L$ & Behavior \\
    \midrule
    Escape Room & Plain PPO & $-0.25 \pm 0.39$ & $4.95 \pm 0.08$ & $-0.25 \pm 0.39$ & no reliable escape \\
    & \alg{TR-EI} (C) & $\mathbf{17.00 \pm 0.00}$ & $6.96 \pm 0.00$ & $-1.00 \pm 0.00$ & concentrated exits, SD 0.49 \\
    & \alg{SRA-EI} & $15.75 \pm 2.95$ & $6.78 \pm 0.42$ & $-1.00 \pm 0.00$ & concentrated exits, SD 0.48 \\
    & \alg{SRR-IA} + Standing (NW) & $13.08 \pm 0.28$ & $\mathbf{7.50 \pm 0.03}$ & $\mathbf{+1.27 \pm 0.12}$ & criterion 8/8, SD 0.13 \\
    \midrule
    Clean Up & Plain PPO & $248 \pm 198$ & $37.0 \pm 29.8$ & $5.4 \pm 4.8$ & -- \\
    & \alg{TR-SVO} (C) & $1{,}405 \pm 316$ & $64.6 \pm 37.8$ & $5.3 \pm 12.5$ & cleaning share $0.86 \pm 0.16$ \\
    & \alg{TR-IA} & $964 \pm 107$ & $178.1 \pm 20.6$ & $95.9 \pm 18.0$ & -- \\
    & \alg{SRA-EI} & $1{,}286 \pm 141$ & $214.4 \pm 15.3$ & $83.1 \pm 16.6$ & cleaning share $0.89 \pm 0.12$ \\
    & \alg{SRR-EI} & $\mathbf{1{,}512 \pm 140}$ & $84.5 \pm 29.8$ & $0$ & one or two agents only clean \\
    & \alg{SRR-IA+V} (NW) & $1{,}353 \pm 93$ & $\mathbf{269.9 \pm 18.8}$ & $\mathbf{251.5 \pm 23.4}$ & cleaning share $0.51 \pm 0.12$ \\
    \bottomrule
  \end{tabular}
\end{table*}

\subsection{Production Is Not Cooperation}
\label{app:nw-curves}

\begin{figure*}[tp]
  \centering
  \includegraphics[width=\textwidth]{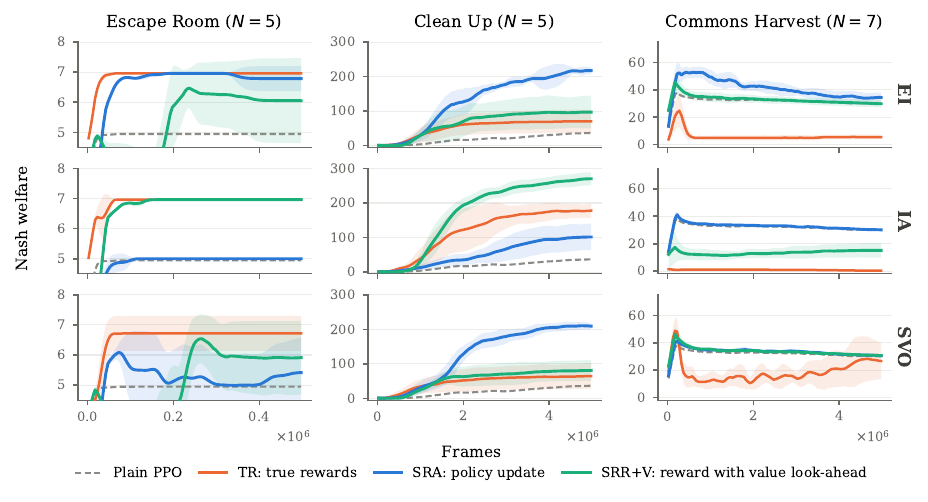}
  \caption{Nash welfare over training for the runs of Figure.3 in the main paper. Columns are games and rows are social
  operators; every learner is at its collective-return selection. Lines are
  means over eight held-out seeds and shaded regions are pointwise 95\%
  Student-$t$ intervals, with a five-bin moving average.}
  \label{fig:nw-curves}
\end{figure*}

\begin{figure*}[tp]
  \centering
  \includegraphics[width=\textwidth]{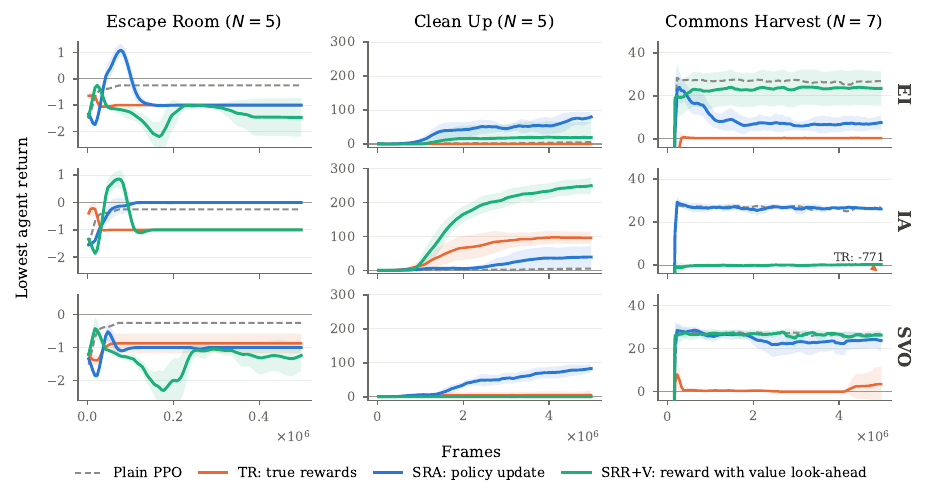}
  \caption{Lowest agent return over training for the runs, with the same layout and intervals as
  Figure~\ref{fig:nw-curves}. The grey line marks zero. Commons Harvest
  \alg{TR-IA} falls below the plotted range, to a final mean of $-771$,
  and is marked at the panel edge.}
  \label{fig:lowest-curves}
\end{figure*}

Collective return says how much the group produces. Figures~\ref{fig:nw-curves} and~\ref{fig:lowest-curves} show the Nash welfare and the lowest agent return of the same runs, in the same layout, and they change the ranking of the learners.

In Clean Up, the true-reward learners lead on collective return under EI and
SVO, but in Nash welfare they fall to the bottom of their panels, near 60,
while \alg{SRA-EI} and \alg{SRA-SVO} climb to about 215 and 210. The lowest
agent return separates them further: the true-reward EI and SVO learners
leave their poorest agent at zero for the whole run, while \alg{SRA} lifts
it to about 83. In Commons Harvest the reversal is starker. \alg{TR-EI}'s
Nash welfare falls to about 6 and its poorest agent to under 1, below plain
PPO on both, while \alg{SRA-EI} stays near 36. The advantage of
observing the true rewards lies in production, and it is paid for by the
agents who end with nothing.

Escape Room shows a ceiling rather than a separation. Every learner that
escapes reaches a Nash welfare near 7 and a lowest agent return of $-1$,
because whoever pulls the lever loses a point in every episode. No learner
without standing avoids this, which is what Appendix~\ref{app:standing}
addresses.

\subsection{Escape Room at Every Group Size}
\label{app:escape}

Table~\ref{tab:app-escape-efficiency} asks whether the Escape Room result
depends on the group size. At every size from two to twelve agents, at
least one self-referenced variant reaches the optimum, so the answer is no:
the game stays solvable without the others' rewards. The last column shows
how narrow the path is. Seven of sixteen variants reach the optimum with
two agents, but only two or three from five agents on. Self-referenced estimates make the game solvable at every size, but a configuration that
works at one size cannot be assumed to work at another.

\begin{table}[tb]
  \centering
  \caption{Escape Room: collective return as a fraction of the optimum,
  held-out seeds. ``Best self-ref.'' is the best of 16 self-referenced
  variants (four operators $\times$ four placements) selected on tuning
  seeds; the last column counts how many of the 16 reach 90\% of the
  optimum.}
  \label{tab:app-escape-efficiency}
  \centering
  \small
  \begin{tabular}{@{}rrrrlrr@{}}
    \toprule
    $N$ & $M$ & PPO & \alg{TR} & Best self-ref. & \alg{SRA-EI} & $\ge 90\%$ \\
    \midrule
     2 & 1 & 0.00 & 1.00 & 1.00 (\alg{SRR-EI}) & 1.00 & 7/16 \\
     3 & 2 & 0.00 & 1.00 & 1.00 (\alg{SRR-IA}) & 0.69 & 3/16 \\
     5 & 3 & $-$0.01 & 1.00 & 1.00 (\alg{SRR-IA}) & 0.93 & 3/16 \\
     7 & 4 & 0.00 & 1.00 & 1.00 (\alg{SRR-IA}) & 0.82 & 2/16 \\
    10 & 5 & 0.00 & 1.00 & 1.00 (\alg{SRR-IA}) & 0.69 & 2/16 \\
    12 & 6 & 0.00 & 1.00 & 1.00 (\alg{SRR-IA}) & 0.00 & 2/16 \\
    \bottomrule
  \end{tabular}
\end{table}

\section{Where Should the Self-Referenced Estimate Enter?}
\label{app:res-placement}

The same self-referenced estimate can enter learning in the reward or in the policy update, and the two places suit different preferences. A preference that compares agents belongs in the reward, with the look-ahead; a benevolent one is safer in the policy update. Two experiments support this division. The first runs the reward placement with and without the look-ahead in every game and under every operator. The second follows both placements as the group grows.

\subsection{What the Look-Ahead Does}
\label{app:lookahead}

\begin{figure*}[tp]
  \centering
  \includegraphics[width=\textwidth]{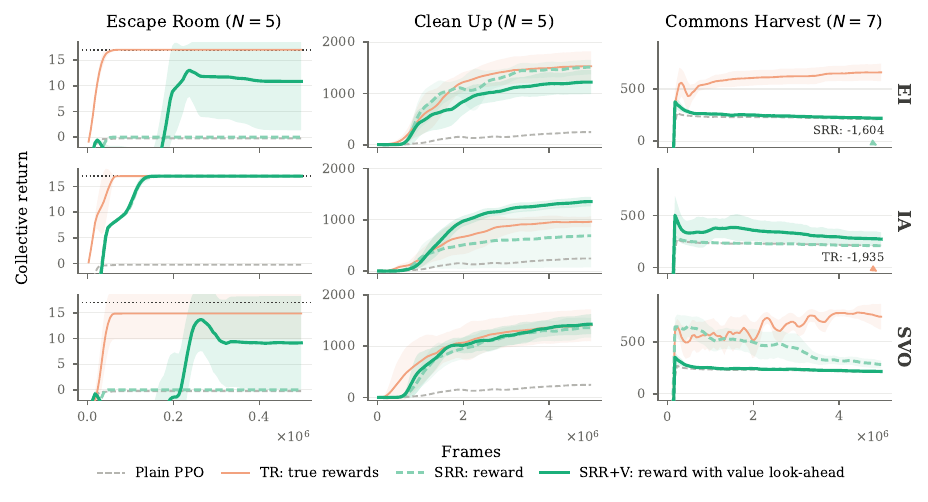}
  \caption{The value look-ahead. Collective return over training for the
  reward placement without the look-ahead, \alg{SRR}, and with it,
  \alg{SRR+V}; plain PPO and the true-reward learner are drawn faintly for
  reference. Every learner is at its collective-return selection; lines are
  means over eight held-out seeds and shaded regions, drawn for every learner
  except plain PPO, are pointwise 95\% Student-$t$ intervals; where all
  seeds agree, as for the true-reward learners at the Escape Room optimum,
  the interval has no width. Learners below
  the plotted range are marked at the panel edge with their final means.}
  \label{fig:placement-curves}
\end{figure*}

Figure~\ref{fig:placement-curves} runs the reward placement with and without the look-ahead in every game and under every operator. The look-ahead adds the critic's value of the other agent's next observation to its self-referenced trace, Eq.7, so an agent is judged by what it has earned and by what it is about to earn.

The look-ahead does two different jobs. In Escape Room under EI and SVO it gets learning started: without it the dashed line stays at zero for the whole run, because agents that never pull the lever never reach the door, their reward model is fitted on transitions with no reward in them, and the social term is zero everywhere. The critic already tells positions apart, and that is enough to push the agents off the selfish solution. Under inequity aversion it prevents a misjudgement: a trace makes an agent that is cleaning look far behind, and an agent averse to being ahead stops eating. Crediting the cleaner with the apples it can expect removes most of that apparent gap, and the collective return in Clean Up doubles.

The same mechanism has a cost when the preference only adds outcomes. Under
EI in Clean Up, \alg{SRR-EI+V} produces about 300 apples less than
\alg{SRR-EI}: crediting the others with what they are about to earn pays
the agent for a benefit not yet produced. In Commons Harvest the look-ahead
does not rescue EI or SVO either; tuning selects the smallest weight we
tried, and larger weights stop the agents harvesting. The look-ahead helps a
preference that compares agents and can hurt one that adds them up.

\subsection{What Happens as the Group Grows}
\label{app:by-n}
\label{sec:results-n}

\begin{figure*}[tp]
  \centering
  \includegraphics[width=\textwidth]{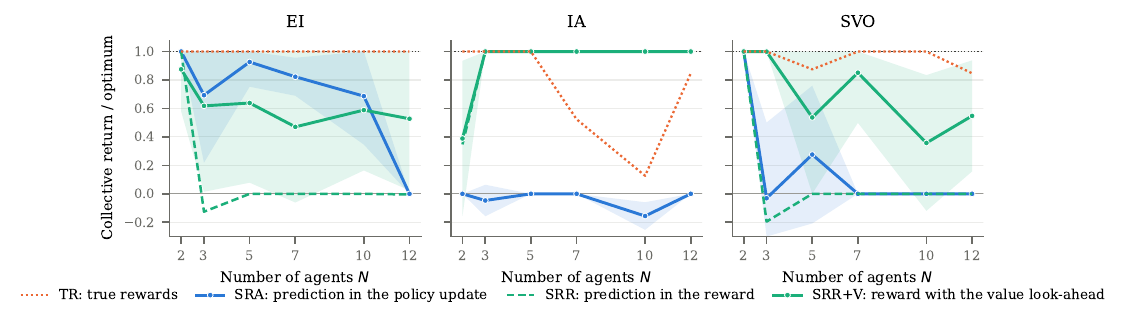}
  \caption{Where the self-referenced estimate enters, by group size. Escape Room
  collective return divided by the optimum for each operator, with the estimate in the policy update, \alg{SRA}, in the reward, \alg{SRR}, and
  in the reward with the value look-ahead, \alg{SRR+V}, against the
  true-reward learner. Every cell is tuned separately at its $N$ on tuning
  seeds; points are means over eight held-out seeds and shaded regions,
  drawn for \alg{SRA} and \alg{SRR+V}, are 95\% Student-$t$ intervals.}
  \label{fig:placement}
\end{figure*}

\begin{figure*}[tp]
  \centering
  \includegraphics[width=\textwidth]{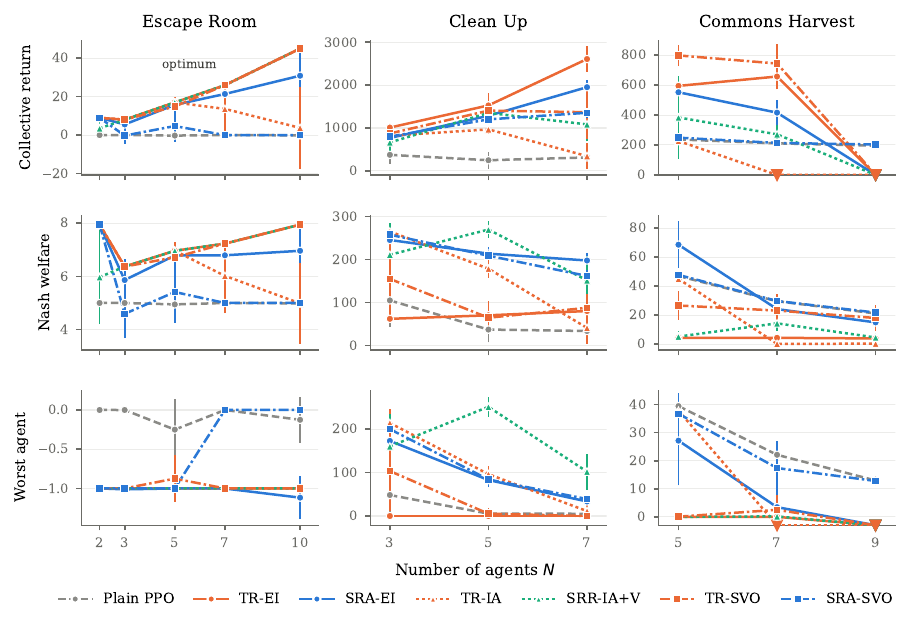}
  \caption{Final collective return, Nash welfare, and lowest agent return
  against the number of agents $N$ for plain PPO, \alg{TR-EI}, \alg{SRA-EI},
  \alg{TR-IA}, \alg{SRR-IA+V}, \alg{TR-SVO}, and \alg{SRA-SVO}, each at its
  collective-return selection for that $N$; means over eight held-out seeds,
  bars are 95\% $t$-intervals. The Escape Room optimum grows with $N$ (dotted
  line). Downward triangles mark means below the plotted range: in Commons
  Harvest, \alg{TR-IA} at $N=7$ ends at $-1{,}935$, and at $N=9$ \alg{TR-IA},
  \alg{SRR-IA+V}, \alg{SRA-EI}, \alg{TR-EI}, and \alg{TR-SVO} end at
  $-12{,}818$, $-5{,}735$, $-3{,}770$, $-2{,}268$, and $-1{,}720$. Commons
  Harvest Nash welfare and lowest agent return here are the per-step logged values,
  which differ slightly from the values computed from per-agent means in the other tables.}
  \label{fig:by-n}
\end{figure*}

Figure~\ref{fig:placement} repeats the comparison of placements in Escape
Room for two to twelve agents, each cell tuned at its own size, and
Figure~\ref{fig:by-n} follows the main learners in all three games and on
all three measures. The two methods fail in different ways as the group
grows.

The policy update dilutes. Its coefficient averages the scores of the
$N-1$ other agents, Eq.12, so the part that
responds to any one agent's action shrinks as $1/(N-1)$. \alg{SRA-EI} reaches
at least two thirds of the optimum up to ten agents and none of it with
twelve, and it
falls further behind \alg{TR-EI} as $N$ grows in Escape Room, where it
reaches 31 against 45 with ten agents, and in Commons Harvest. The reward
placement does not dilute: under inequity aversion \alg{SRR} and
\alg{SRR+V} sit at the optimum from three agents up to twelve, while
\alg{SRA-IA} stays at zero at every size, because a comparison that only
reweights scores does nothing at any $N$.

Inequity aversion also behaves differently with true rewards and with self-referenced estimates. With true rewards it declines as the group grows, \alg{TR-IA}
falling to 3.8 in Escape Room with ten agents and from 820 to 345 apples in
Clean Up between three and seven agents, and it collapses in Commons
Harvest. With self-referenced estimates and the look-ahead it holds:
\alg{SRR-IA+V} produces most with five agents, gives its poorest agent more
than any other learner shown with five and seven, and collapses only with nine agents in Commons
Harvest, where every learner with a social term except \alg{SRA-SVO}
collapses. A larger group gives an inequity-averse agent more pairs to be
behind in, and the true rewards make each of those comparisons sharper than
it is. We have not isolated this mechanism.

\section{Who Pays for Cooperation?}
\label{app:res-who-pays}

A high collective return can hide who receives it and who does the work. In Clean Up the most productive learners split into eaters and cleaners, while \alg{SRR-IA+V} shares the work. This appendix shows that the pattern holds for all nine learners. It also appears, in its own form, in the other two games, and it survives a change of the selection metric.

\subsection{All Nine Learners in Clean Up}
\label{app:agents}

\begin{figure*}[tp]
  \centering
  \includegraphics[width=\textwidth]{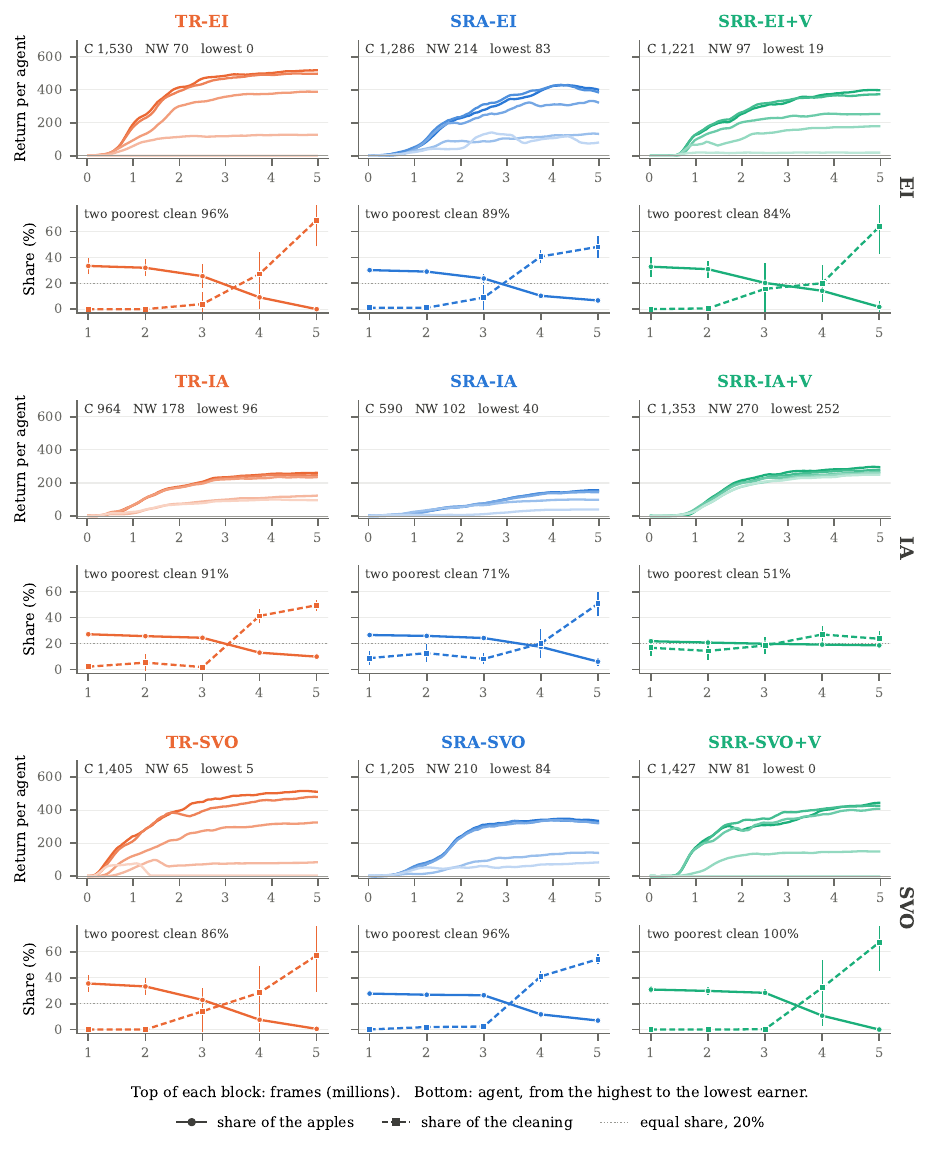}
\caption{Who receives the return and who does the work for all nine Clean Up learners with five agents. One block per operator; columns are the true-reward learner, \alg{SRA}, and \alg{SRR+V}, each at its collective-return selection. Top row of each block: every agent's return over training, agents ranked within each seed by their final return and averaged by rank over the eight held-out seeds, darkest the richest. Bottom row: each agent's final share of the apples, solid, and of the cleaning, dashed, with 95\% Student-$t$ intervals; the dotted line is the equal share of 20\%.}
  \label{fig:app-agents-cu-extended}
\end{figure*}

Figure~\ref{fig:app-agents-cu-extended} shows all nine Clean Up learners, one block per operator, with every agent's
return over training on top and the shares of the apples and the cleaning
below. Read down a column, the two rows tell the same story from two sides:
an agent whose return line stays near zero is, in the row beneath, the one
doing most of the cleaning.

The split into eaters and cleaners is the rule, not the exception. In eight
of the nine learners the line of the apples and the line of the cleaning
cross, and the two poorest agents do between 71\% and 100\% of the cleaning
while the richest does almost none. It forms early and does not change: in
the four learners that leave their poorest agent near zero, \alg{TR-EI},
\alg{TR-SVO}, \alg{SRR-EI+V}, and \alg{SRR-SVO+V}, that agent's line
flattens within the first two million frames and never rises. Where the estimate enters matters more than the estimate itself. \alg{SRR-EI+V} and \alg{SRR-SVO+V} use the same self-referenced estimates as \alg{SRA-EI} and \alg{SRA-SVO}, yet divide the
return as unequally as the true-reward learners. The one learner that breaks
the pattern is \alg{SRR-IA+V}: its five return lines climb together, and
both of its share lines stay near 20\% for every agent, so its high Nash
welfare is not an artefact of everyone earning equally little.

The Nash welfare puts numbers on this. For every operator our learner's welfare is far above that of the true-reward learner with the same operator: 214 against 70 under EI, 270 against 178 under inequity aversion, and 210 against 65 under SVO. The gap comes from the poorest agents, not from production, since the collective returns differ by a sixth at most and ours is higher under inequity aversion; a geometric mean is pulled down by the agent with the least, and the true-reward EI and SVO learners leave that agent with almost nothing. The reward placement with EI and SVO is the cautionary case: it uses the same self-referenced estimates as \alg{SRA}, yet its welfare is no better than the true-reward learner's, 97 and 81, because its two poorest agents do 84\% and 100\% of the cleaning.

Tables~\ref{tab:app-cleanup-n3} and~\ref{tab:app-cleanup-n7} show that the pattern does not depend on the group size. With seven agents, under
\alg{SRA-EI} the two poorest agents do $86 \pm 9$\% of the cleaning and
receive 5\% of the apples, while under \alg{SRR-IA} they do $34 \pm 7$\%
and receive 24\%.

\begin{table*}[tp]
  \centering
  \caption{Clean Up, three agents, eight held-out seeds (95\% CIs). IA
  settings are $\alpha/\beta$; SVO uses $\phi = \pi/3$. The last column
  lists the mean return of the highest through the lowest earner.}
  \label{tab:app-cleanup-n3}
  \centering
  \small
  \begin{tabular}{@{}llrrrl@{}}
    \toprule
    Method & Setting & $C$ & $\NW$ & $L$ & Returns \\
    \midrule
    Plain PPO & -- & $373 \pm 206$ & $104.9 \pm 61.8$ & $48.3 \pm 45.2$ & 172 / 153 / 48 \\
    \alg{SRA-EI} (NW) & $\alpha{=}2$ & $768 \pm 69$ & $245.6 \pm 24.5$ & $172.8 \pm 27.9$ & 335 / 260 / 173 \\
    \alg{SRA-SVO} & $\alpha{=}7$ & $807 \pm 36$ & $258.9 \pm 13.5$ & $196.1 \pm 23.3$ & 369 / 242 / 196 \\
    \alg{SRA-IA} & $0/1$ & $689 \pm 32$ & $228.7 \pm 10.7$ & $208.6 \pm 13.8$ & 257 / 223 / 209 \\
    \alg{SRR-IA+V} & $0/0.03$ & $664 \pm 213$ & $211.2 \pm 74.1$ & $160.3 \pm 75.1$ & 296 / 207 / 160 \\
    \alg{TR-IA} & $0/0.03$ & $820 \pm 37$ & $\mathbf{265.4 \pm 16.1}$ & $\mathbf{215.7 \pm 31.9}$ & 361 / 243 / 216 \\
    \alg{TR-EI} (C) & $\alpha{=}0.01$ & $\mathbf{1{,}010 \pm 19}$ & $62.0 \pm 0.9$ & $0$ & 585 / 425 / 0 \\
    \bottomrule
  \end{tabular}
\end{table*}

\begin{table*}[tp]
  \centering
  \caption{Clean Up, seven agents, eight held-out seeds (7--14; 95\% CIs).
  Each learner at the setting selected on the three tuning seeds, by the metric in parentheses. The last column lists the mean return of the highest
  through the lowest earner.}
  \label{tab:app-cleanup-n7}
  \centering
  \small
    \begin{tabular}{@{}llrrrl@{}}
    \toprule
    Method & Setting & $C$ & $\NW$ & $L$ & Returns, high $\to$ low \\
    \midrule
    Plain PPO & -- & $313 \pm 193$ & $33.7 \pm 24.0$ & $5.2 \pm 8.2$ & 70 / 60 / 51 / 49 / 47 / 31 / 5 \\
    \alg{SRA-EI} (NW) & $\alpha{=}3$ & $1{,}953 \pm 174$ & $\mathbf{198.5 \pm 18.2}$ & $34.2 \pm 13.7$ & 426 / 409 / 383 / 352 / 285 / 64 / 34 \\
    \alg{SRA-SVO} (NW) & $\alpha{=}5$ & $1{,}356 \pm 111$ & $161.8 \pm 13.9$ & $39.0 \pm 9.8$ & 266 / 252 / 244 / 239 / 231 / 85 / 39 \\
    \alg{SRR-IA} (NW) & $0.003/0.1$ & $1{,}069 \pm 116$ & $151.1 \pm 16.2$ & $\mathbf{116.3 \pm 17.8}$ & 178 / 166 / 162 / 155 / 150 / 142 / 116 \\
    \alg{SRR-IA+V} (NW) & $0.003/0.1$ & $1{,}006 \pm 133$ & $140.9 \pm 19.8$ & $100.3 \pm 26.9$ & 181 / 166 / 154 / 143 / 133 / 129 / 100 \\
    \alg{TR-IA} (NW) & $0/0.03$ & $345 \pm 296$ & $40.9 \pm 37.5$ & $11.4 \pm 14.8$ & 76 / 65 / 62 / 55 / 52 / 23 / 11 \\
    \alg{TR-EI} (C) & $\alpha{=}0.03$ & $\mathbf{2{,}609 \pm 298}$ & $80.1 \pm 15.8$ & $0$ & 568 / 555 / 545 / 541 / 401 / 0 / 0 \\
    \alg{TR-SVO} (C) & $\alpha{=}0.01$ & $1{,}369 \pm 623$ & $87.6 \pm 32.6$ & $0$ & 276 / 265 / 251 / 245 / 239 / 92 / 0 \\
    \bottomrule
  \end{tabular}
\end{table*}

\subsection{The Same Tension in the Other Games}
\label{app:agents-other}

\begin{figure*}[tp]
  \centering
  \includegraphics[width=\textwidth]{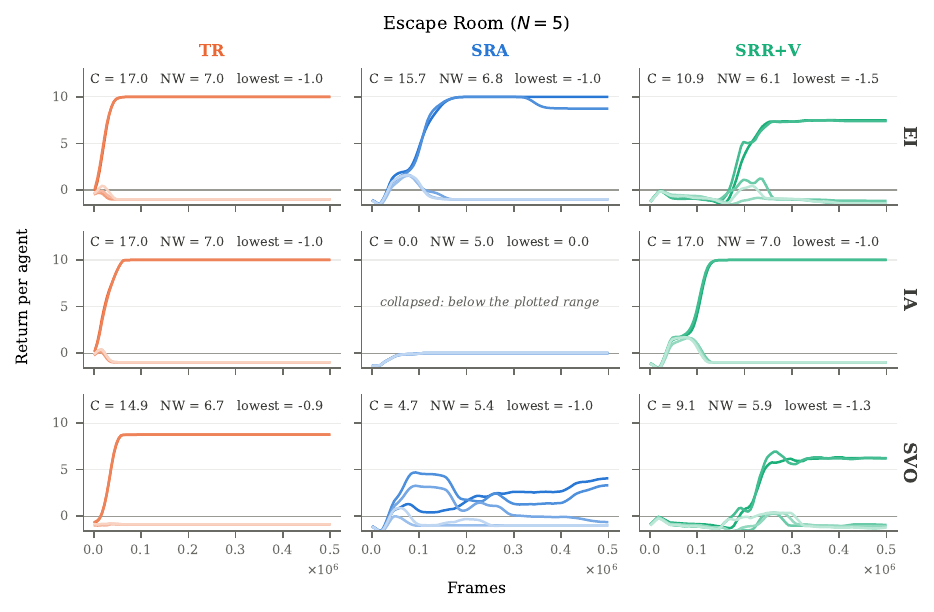}
  \caption{Who receives the return in Escape Room with five agents, as in
  Figure~\ref{fig:app-agents-cu-extended}. Nash welfare uses the offset $c = 5$ of Eq.~\eqref{eq:collective-lowest}.}
  \label{fig:agents-er}
\end{figure*}

\begin{figure*}[tp]
  \centering
  \includegraphics[width=\textwidth]{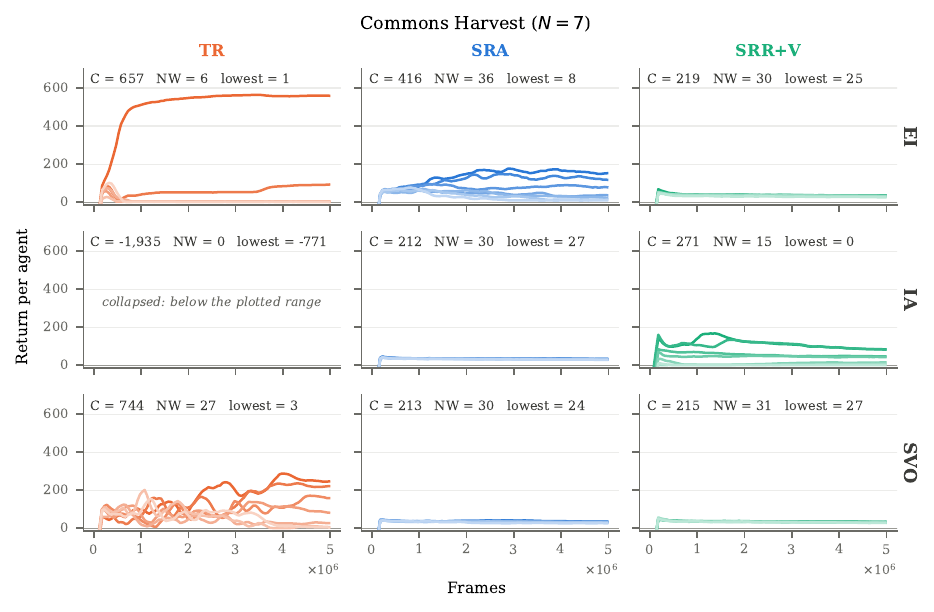}
  \caption{Who receives the return in Commons Harvest with seven agents, as
  in Figure~\ref{fig:app-agents-cu-extended}. \alg{TR-IA} collapses and lies below the
  plotted range.}
  \label{fig:agents-hv}
\end{figure*}

Figures~\ref{fig:agents-er} and~\ref{fig:agents-hv} show every agent's
return in Escape Room and Commons Harvest. The tension between producing
and sharing appears in both, in a form set by the game.

In Escape Room it is built in. Every learner that escapes produces the same
two-tier outcome, two agents at $+10$ and three at $-1$, because three of
five must hold the lever while the other two walk out. That this is the same
under \alg{TR-EI}, \alg{SRA-EI}, and \alg{SRR-IA+V} shows it is a property
of the game, not of any learner, and collective return cannot tell a group
that rotates the lever from one that does not, since every episode pays the
same whoever pulls. This is why the game needs standing,
Appendix~\ref{app:standing}.

In Commons Harvest it takes the form of a monopoly. Under \alg{TR-EI} one
agent climbs to about 560 of the group's 657 apples while the rest stay near
the floor, so the game's highest collective return goes almost entirely to
one agent. \alg{SRA-EI} produces less, 416, but its lines stay together and
its poorest agent ends at 8 against under 1. The Nash welfare tells the same story more sharply: \alg{TR-EI}'s falls to 6, below even plain PPO's 30, while \alg{SRA-EI}'s is 36. Welfare in this game needs care, though. The self-scaling control, which ignores the others, reaches the highest welfare of all, so in Commons Harvest a high welfare is not by itself evidence of a social preference.

\subsection{Selecting on Nash Welfare}
\label{app:pick-plane}

\begin{figure*}[tp]
  \centering
  \includegraphics[width=\textwidth]{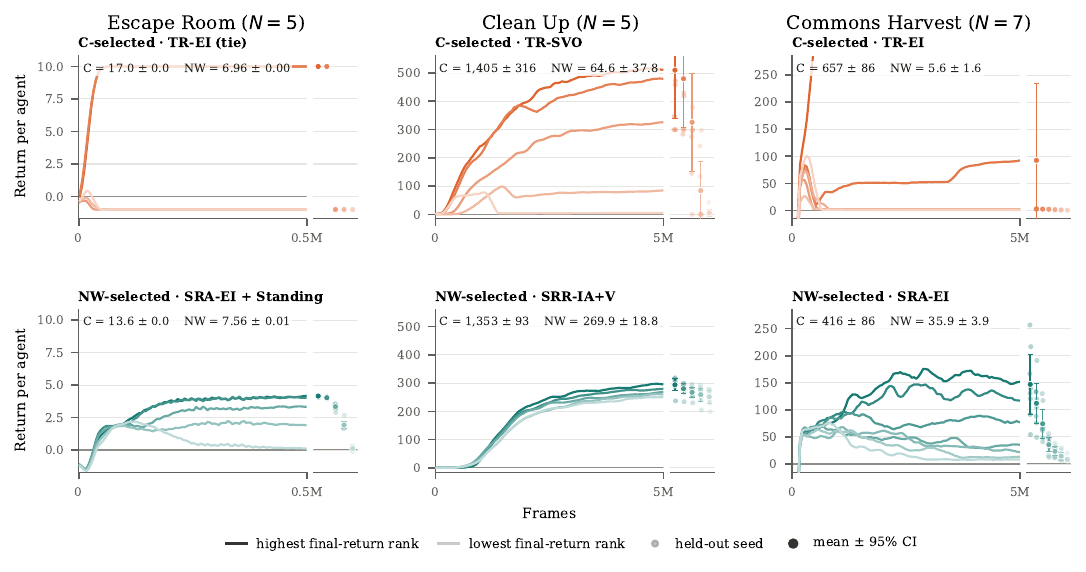}
  \caption{Agent outcomes under the two tuning objectives. Columns are the
  three games; rows are the configurations selected on tuning seeds by
  collective return (top) and Nash welfare (bottom), without the self-scaling controls.
  Each large panel shows the return of every agent over training:
  agents are ordered within each seed by their final-window return and
  then averaged by rank across the eight held-out seeds (darkest: highest
  final-return rank). The narrow panel shows all eight final seed values and
  the rank mean $\pm$ 95\% Student-$t$ interval. Insets give the final $C$ and $\NW$. Ranked curves describe how the return is divided; they do not follow one agent across seeds.}
  \label{fig:selection-agents}
\end{figure*}

A natural objection is that our learners only look fairer because of how we
selected them. Figure~\ref{fig:selection-agents} answers it by taking, in
each game, the one configuration that each metric selects among all social
learners, and plotting every agent's return. The two rows differ in shape,
not only in level: the top row, selected on collective return, has a leader
and a floor, and the bottom row, selected on Nash welfare, has lines that
stay together. In Clean Up the collective-return choice, \alg{TR-SVO},
produces 1{,}405 apples with a Nash welfare of 65, and the Nash-welfare
choice, \alg{SRR-IA+V}, produces 1{,}353 with a welfare of 270, so almost
the same harvest is divided very differently. The tuning landscape of
Figure~\ref{fig:app-cleanup-landscape} shows that this is not one lucky
pick: at every group size, the collective-return choice is a true-reward
learner and the Nash-welfare choice is one of ours.

For a single learner the two metrics usually agree, and where they differ
the effect is not consistent. In Escape Room with seven agents, Nash welfare
selects \alg{SRR-IA} with standing at $\kappa = 2$, which reaches 72\% of
the optimum, over $\kappa = 1$, which produces more but leaves the poorest
agent at a loss. In Clean Up with five agents, the \alg{TR-SVO} setting
chosen on Nash welfare produced more apples on the held-out seeds,
$1{,}710 \pm 499$, than the one chosen on collective return,
$1{,}405 \pm 316$, while still leaving one agent with nothing; we report the
setting chosen on the tuning seeds and do not reselect. In Commons Harvest
the Nash-welfare settings make the \alg{SRR} learners equal by reducing
their production to that of plain PPO, which is the failure mode of the
metric. Neither metric is enough alone, which is why we report both and the
lowest agent return as well.

\section{Standing in the Repeated Game}
\label{app:standing}

In Escape Room some agents must pay for the lever in every episode, so a fair division is possible only across the episodes of a run. Standing, achieves it: fixed roles become shared ones, at the cost of about a quarter of the optimum.
This appendix explains how standing is computed, shows that the result holds
on every seed, traces the lost quarter to coordination, and explains why we
do not use standing in Clean Up.

\subsection{How Standing Is Computed}
\label{app:standing-def}

Each observer $i$ keeps, for every agent $j$, a running average $G_{ij}$ of
$j$'s episode return as $i$ estimates it from self-reference, and of its own true return when
$j = i$. After episode $k$,
\begin{equation}
  \hat R_{ij}^{k}=\sum_{t\in k}\hat r_{ij}^t,
  \qquad
  \eta_k=\max(1/k,1/H),
  \label{eq:ssa-standing-r}
\end{equation}
\begin{equation}
  G_{ij}^{k}=(1-\eta_k)G_{ij}^{k-1}+\eta_k\hat R_{ij}^{k},
  \label{eq:ssa-standing}
\end{equation}
a plain mean over the first $H$ episodes and an exponential moving average
after that. The mean at the start keeps early episodes from being dominated
by the latest one, and the moving average afterwards lets the ledger follow
a change in behaviour. We use $H = 100$ episodes in Escape Room and
$H \in \{20, 30\}$ in Clean Up, whose 1{,}000-step episodes give only a few
hundred episodes per environment copy. Every copy of the environment keeps
its own ledger, and each entry is updated when that agent's episode ends.

From the ledger, $i$ measures how far it has been ahead of the others and
how far behind them,
\begin{align}
 a_i &= \frac{1}{N-1}\sum_{j\neq i}[G_{ii}-G_{ij}]_+, \qquad
 b_i = \frac{1}{N-1}\sum_{j\neq i}[G_{ij}-G_{ii}]_+,
 \nonumber\\
 \mathrm{ahead}_i &= \frac{a_i}{a_i + c_i}, \qquad
 \mathrm{behind}_i = \frac{b_i}{b_i + c_i},
 \label{eq:app-standing-gaps}
\end{align}
with the scale $c_i = |G_{ii}| + (N-1)^{-1}\sum_{j\neq i}|G_{ij}|$ plus a
small constant. Dividing by a scale built from the returns themselves lets
one gate strength $\kappa$ work in games whose returns differ by two orders
of magnitude: both fractions lie in $[0,1)$ and do not change when every
return is multiplied by the same constant. Both are zero when $i$'s history
is level with the others', so the gate leaves a fair division alone and
only separates ``always me'' from ``each of us in turn''.

The gate is off at the start of training, and this matters. It waits for
the reward-model warm-up $W$ and a further gate warm-up, and then acts only
once some agent's average exceeds a threshold, two per episode in every
Escape Room run. Before that the group has nothing to share: a gate applied
from the start discounts the first agent that earns anything, and in our
runs it prevents cooperation from forming at all. The threshold is tested on
all agents, the observer included, because with three agents and two
pulling the lever only one agent collects, and testing the others alone
would never switch the gate on for the agent that is ahead.

\subsection{The Result Holds on Every Seed}
\label{app:standing-er}

\begin{figure*}[tp]
  \centering
  \includegraphics[width=\textwidth]{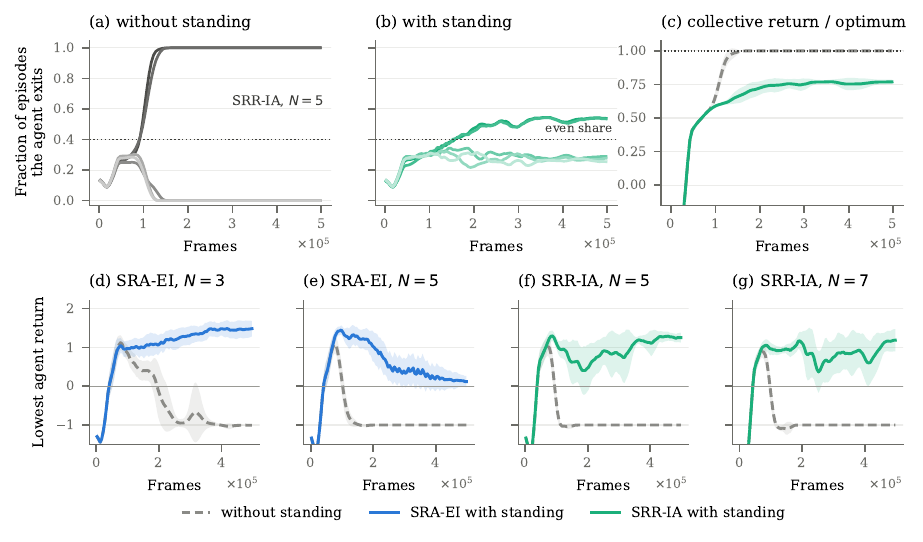}
  \caption{Standing in the repeated game, Escape Room. (a, b) \alg{SRR-IA}
  with five agents, without and with standing, the only difference between
  the two: the fraction of episodes in which each agent exits, ranked
  within each seed by its final exit rate and averaged by rank over the
  eight held-out seeds; the dotted line is the even share, two exits in
  five. (c) Collective return divided by the optimum for the same two runs.
  (d--g) Lowest agent return over training without and with standing for
  the four exact pairs. Means over seeds 11--18 with pointwise 95\%
  Student-$t$ intervals. Figure~\ref{fig:app-ledger-seeds} shows every seed.}
  \label{fig:ledger}
\end{figure*}

Figure~\ref{fig:ledger} shows the four pairs of configurations that differ only in standing: \alg{SRA-EI} with three and five agents and \alg{SRR-IA} with five and seven. Panels (a) and (b) are the result. Without standing the
exit rates split within the first 150 thousand frames and stay split: two
agents exit in every episode and three in none, on every seed. With
standing no line sits at zero, every agent exits in at least a quarter of
the episodes, and the lines gather around the even share of two in five.
Panel (c) gives the price, a collective return that settles at about three
quarters of the optimum, and panels (d)--(g) show the same trade in all four
pairs.

\begin{figure*}[tp]
  \centering
  \includegraphics[width=\textwidth]{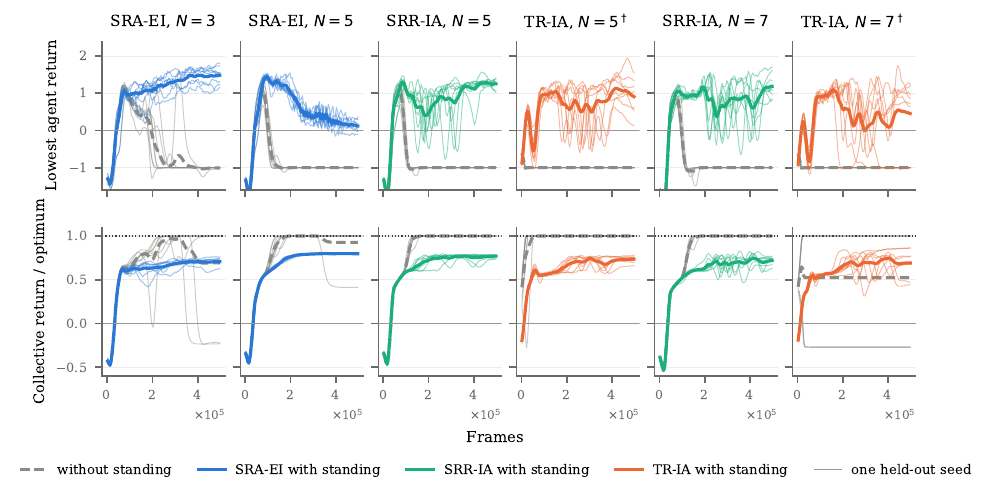}
  \caption{Every held-out seed of the standing pairs in Escape Room over
  training: the lowest agent return, top, and the collective return over
  the optimum, bottom, without standing in grey and with it in colour. Thin
  lines are single seeds, seeds 11--18, and bold lines their mean. Without
  standing, every seed settles at a loss of one point for its worst-off
  agent, and most groups reach the optimum; with standing, the
  worst-off agent stays above zero on almost every seed, and the group
  settles at 70 to 80\% of the optimum. $^\dagger$The two \alg{TR-IA} pairs
  also differ in their PPO settings.}
  \label{fig:app-ledger-seeds}
\end{figure*}

A mean can hide a mixture of outcomes, so Figure~\ref{fig:app-ledger-seeds}
shows every held-out seed. Without standing every seed settles at a loss of
one point for its worst-off agent; with standing the worst-off agent is
above zero on almost every seed. Table~\ref{tab:app-escape-balance} counts
them: standing takes every pair from no shared seed to six or eight of
eight. A larger sample confirms the two \alg{SRA-EI} pairs. Over twenty
seeds the group with three agents reaches $5.62 \pm 0.37$ of an optimum of
8, its lowest agent gains $1.48 \pm 0.30$ per episode, and the agents exit
on 27, 28, and 30\% of episodes, close to the 33\% of an even split. With
five agents the group reaches $13.56 \pm 0.10$ of 17 and exit rates lie
between 35 and 41\%, against 40\% for an even split. These are frequencies
over many episodes, not an order in which the agents take turns.

\begin{table*}[tp]
  \centering
  \caption{The four exact pairs of Figure~\ref{fig:ledger} and the two descriptive \alg{TR-IA} pairs,
  held-out seeds 11--18. Values for the with-standing arm are means $\pm$
  95\% Student-$t$ interval half-widths. A seed counts as shared when $L_s>0$ and the exit rates have a standard deviation below 0.20; the last column gives the shared seeds without $\rightarrow$ with standing. Every
  without-standing arm is 0/8. $^\dagger$The \alg{TR-IA} pairs also differ in their PPO settings.}
  \label{tab:app-escape-balance}
  \centering
  \small
    \begin{tabular}{@{}lrrrc@{}}
    \toprule
    ER($N$,$M$), with-standing configuration & $C/C^\star$ & $L_s$ & Exit SD & Shared seeds \\
    \midrule
    (3,2) \alg{SRA-EI}, $\kappa{=}4$ & $0.711\pm0.035$ & $1.514\pm0.220$ & $0.030\pm0.015$ & $0/8\to8/8$ \\
    (5,3) \alg{SRA-EI}, $\kappa{=}4$ & $0.797\pm0.002$ & $0.114\pm0.119$ & $0.129\pm0.003$ & $0/8\to6/8$ \\
    (5,3) \alg{SRR-IA}, $\kappa{=}1$ & $0.769\pm0.016$ & $1.268\pm0.122$ & $0.131\pm0.016$ & $0/8\to8/8$ \\
    (5,3) \alg{TR-IA}, $\kappa{=}1$$^\dagger$ & $0.735\pm0.032$ & $0.942\pm0.403$ & $0.107\pm0.020$ & $0/8\to8/8$ \\
    (7,4) \alg{SRR-IA}, $\kappa{=}2$ & $0.718\pm0.039$ & $1.201\pm0.282$ & $0.104\pm0.040$ & $0/8\to8/8$ \\
    (7,4) \alg{TR-IA}, $\kappa{=}2$$^\dagger$ & $0.691\pm0.134$ & $0.476\pm0.783$ & $0.122\pm0.050$ & $0/8\to6/8$ \\
    \bottomrule
  \end{tabular}
\end{table*}

\subsection{Why Sharing Costs a Quarter of the Optimum}
\label{app:standing-cost}

\begin{figure}[tb]
  \centering
  \includegraphics[width=\columnwidth]{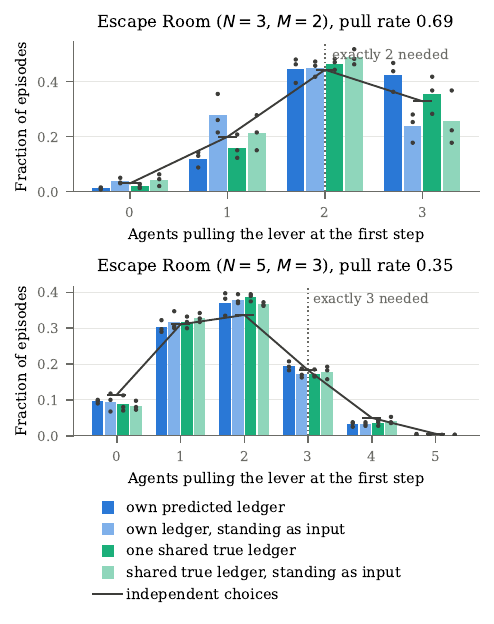}
  \caption{How many agents pull the lever at the first step of an episode
  in Escape Room with standing, over 400 evaluation episodes per run, for the
  four configurations with standing. Bars are means over training seeds and
  dots single seeds. The line is the binomial distribution that independent
  choices with the observed pull rate would give; the episode is solved at
  once only when exactly $M$ agents pull.}
  \label{fig:app-escape-volunteers}
\end{figure}

With standing, every configuration reaches between 69\% and 80\% of the optimum. Escape Room is solved in one step only if exactly the right number of agents pull the lever at once; too few and the door stays shut, too many and
the group pays for nothing. Once every agent sometimes pulls, the question is
whether they manage to pull in the right numbers.
Figure~\ref{fig:app-escape-volunteers} shows that they do not coordinate.
The number of agents that pull at the first step is what independent
choices would give. With five agents and three needed, each agent pulls
first with probability $\bar p = 0.35$, independent choices give exactly
three pullers with probability $\binom{5}{3}\bar p^{3}(1-\bar p)^{2} =
0.18$, and the observed share is 0.18. With three agents and two needed,
$\bar p = 0.69$ gives 0.44 against the observed 0.46. The lost quarter is the
cost of that mismatch.

If the agents knew more, would they coordinate better?
Table~\ref{tab:app-standing-controls} tests three ways of giving them more
information, for the selected \alg{SRA-EI} learner. Training the critic
used in the cross scores on raw returns lowers the return. Giving the policy
its own standing as an input, so that it can act on how far ahead it has
been, changes little. Replacing the private self-referenced ledgers with one ledger of true returns shared by all agents, which removes both the estimation error and the disagreement between ledgers, also changes little.
None of the three recovers the lost quarter, so what is missing is a way to
agree on who pulls this time, not a better estimate of who has been paying.

The standing input is the only control that goes beyond decentralized
execution, because the policy then sees a quantity built from the other
agents' transitions while it acts. It enters the policy and the critic
through a layer initialized to zero,
\begin{equation}
 h \leftarrow h \odot (1 + W_\gamma s_i) + W_\beta s_i,
 \qquad s_i = (\mathrm{ahead}_i, \mathrm{behind}_i),
 \label{eq:app-standing-input}
\end{equation}
so the network is exactly the unconditioned one until the agent has a
standing, and the standing is updated after every step so that the next
action can depend on it. The shared true ledger keeps the reward models and
the self-referenced social term and replaces only the ledger, so it changes two
things at once, and its null result is weaker evidence than it looks.

\begin{table*}[tp]
  \centering
  \caption{Collective return of the selected \alg{SRA-EI} learner in Escape
  Room, without and with each control, and without or with the standing
  gate at $\kappa = 4$; three seeds per cell. The optimum is 8, 17, and 26
  at $N = 3$, 5, and 7.}
  \label{tab:app-standing-controls}
  \centering
  \small
    \begin{tabular}{@{}llrrr@{}}
    \toprule
    Control & Gate & $N = 3$ & $N = 5$ & $N = 7$ \\
    \midrule
    Critic trained on raw returns & off & $8.00 \to 7.44$ & $15.91 \to 13.62$ & $21.60 \to 19.97$ \\
                                  & on  & $5.98 \to 4.67$ & $13.52 \to 13.03$ & -- \\
    Standing as a policy input    & off & $8.00 \to 8.00$ & $15.91 \to 15.99$ & $21.60 \to 21.43$ \\
                                  & on  & $5.98 \to 6.40$ & $13.52 \to 13.53$ & -- \\
    One shared true ledger        & on  & $5.98 \to 5.70$ & $13.52 \to 13.49$ & -- \\
    \quad with standing input     & on  & $6.40 \to 6.17$ & $13.53 \to 13.54$ & -- \\
    \bottomrule
  \end{tabular}
\end{table*}

\subsection{Why Not Clean Up}
\label{app:cleanup}

Standing is a mechanism for a role that can be taken in turns, and Clean Up
does not have one: cleaning is continuous work. The learner that divides
Clean Up best already shares that work, since \alg{SRR-IA+V}'s two poorest
agents do 51\% of the cleaning, Figure~\ref{fig:app-agents-cu-extended}. We
therefore tried standing on the learner that concentrates the work,
\alg{SRA-EI} with five agents, varying its strength, horizon, aggregate,
and source. On the three tuning seeds no setting improved on \alg{SRA-EI}
without standing, which produces 1{,}551 apples with a Nash welfare of 227
and a lowest agent return of 78. The best of the 24 settings reaches a Nash
welfare of 249 and a lowest agent return of 86, within the spread between
seeds, and the four strongest settings lower the collective return to
between 481 and 830 without helping the poorest agent. No setting passed
selection, so none was run on held-out seeds.

\section{What if Agents See Only Part of the Map?}
\label{app:po}

With a $15\times15$ window, agent $i$ must assess $j$'s situation from $j$'s own small window, a part of the map that $i$ may never have seen from there. The two placements then part ways. The policy update keeps its cooperation, while inequity aversion in the reward collapses. The collapse comes from the look-ahead, not from the self-referenced estimates. This appendix gives the evidence for each step and asks whether longer training changes the picture in Commons Harvest.

\subsection{The Policy Update Keeps Cooperating}
\label{app:po-sra}

With feed-forward agents in Clean Up, \alg{SRA-EI} produces about 1{,}280
apples, as with the full view, against about 1{,}900 for \alg{TR-EI}, which
leaves its poorest agent with nothing. With recurrent agents \alg{SRA-EI}
rises to 1{,}725 apples and the highest Nash welfare of all learners, 224,
while \alg{TR-EI} produces 2{,}196 and again leaves its poorest agent at
zero: the same split into eaters and cleaners as with the full view, now
with memory. In Commons Harvest \alg{SRA-EI} produces 646 apples with
feed-forward agents and 482 with recurrent ones, against 187 for plain PPO.
These Harvest runs train for 7 million frames and are retuned, so they are
not directly comparable with the full view. Plain PPO with a GRU fails in
Clean Up at its selected setting, earning three apples, which shows the
risk of selecting on two tuning seeds.

\subsection{Why Inequity Aversion Collapses}
\label{app:po-ia}

\alg{SRR-IA+V} earns nothing in Clean Up with feed-forward agents, recovers
only partly with recurrent ones, to about 400 apples, and stays at plain
PPO's level in Commons Harvest. The self-referenced estimates do not explain this. Their correlation with the true rewards stays between 0.98 and 0.99 with the small window, as high as with the full view. Even the agents that earn almost nothing estimate their few rewards with a mean error of 0.01.

Two controls with feed-forward agents in Clean Up isolate the look-ahead.
\alg{SRR-IA} with the same weights but without the look-ahead produces 111
apples on average instead of none. \alg{TR-IA+V}, which uses the true
rewards together with the look-ahead, produces nothing at all, while
\alg{TR-IA} without it reaches 573 apples at its own selected setting. With true rewards and self-referenced estimates alike, adding the look-ahead under a partial
view destroys cooperation. The look-ahead adds the critic's value of the
other agent's next observation, and under a partial view that observation
shows a different part of the map from anything the critic has learned to
value from there, so the levels that inequity aversion compares are
misjudged and the agent reacts to a gap that is not there.

\subsection{Longer Training in Commons Harvest}
\label{app:po-14m}

\begin{figure*}[tp]
  \centering
  \includegraphics[width=\textwidth]{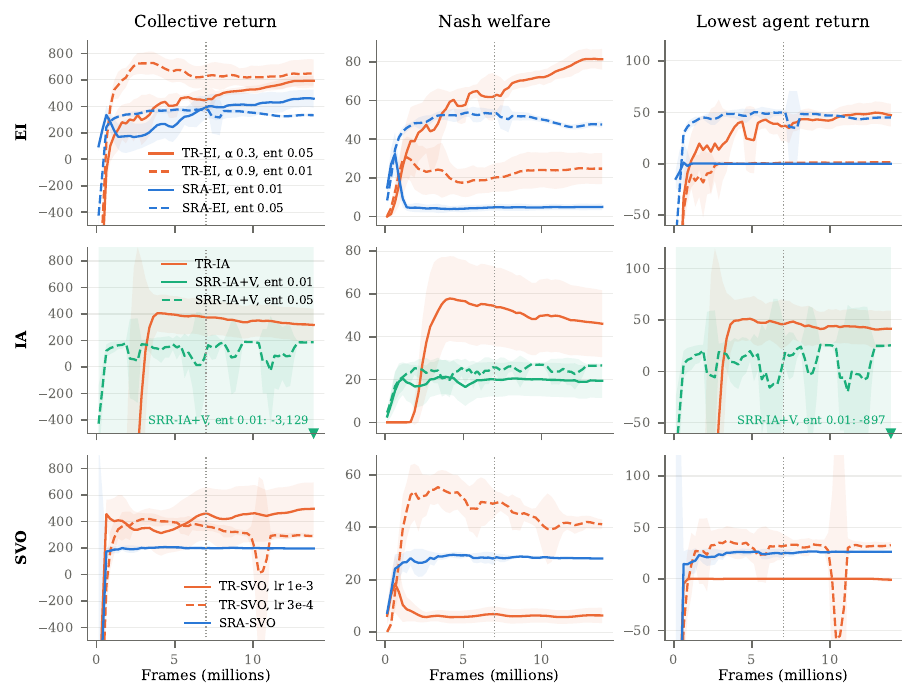}
  \caption{Commons Harvest with recurrent agents and the partial view,
  trained for 14 million frames; the dotted line marks 7 million, the length of the runs in Figure.4 in the main paper. Rows are operators and
  columns the collective return, Nash welfare, and lowest agent return; each
  line is one candidate setting, solid and dashed for the two settings of a
  learner, means over the eight held-out seeds with pointwise 95\%
  Student-$t$ intervals. One \alg{TR-SVO} setting had not finished when the
  figure was made.}
  \label{fig:app-hv-14m}
\end{figure*}

Recurrent agents in Commons Harvest produce less than feed-forward ones, so
we asked whether they were simply undertrained. Figure~\ref{fig:app-hv-14m}
reruns the candidate settings for 14 million frames, twice the length of the
other partial-view runs. Our learners barely change after 7 million frames:
\alg{SRA-EI}, \alg{SRA-SVO}, and \alg{SRR-IA+V} stay close to the level they
had reached, and \alg{SRR-IA+V} with the lower entropy collapses into beam
fights. Only the true-reward EI learner keeps improving, in both collective
return and Nash welfare, so the extra training widens the gap rather than
closing it. The difficulty with recurrent agents in this game is therefore
not a matter of training time. It is also the one setting in which our
fairness advantage disappears: there, the fairer setting of \alg{TR-EI}
leaves its poorest agent better off than any of ours.

\FloatBarrier
\section{Agents That Value Apples Differently}
\label{app:hetero}

\begin{figure}[tb]
  \centering
  \includegraphics[width=\columnwidth]{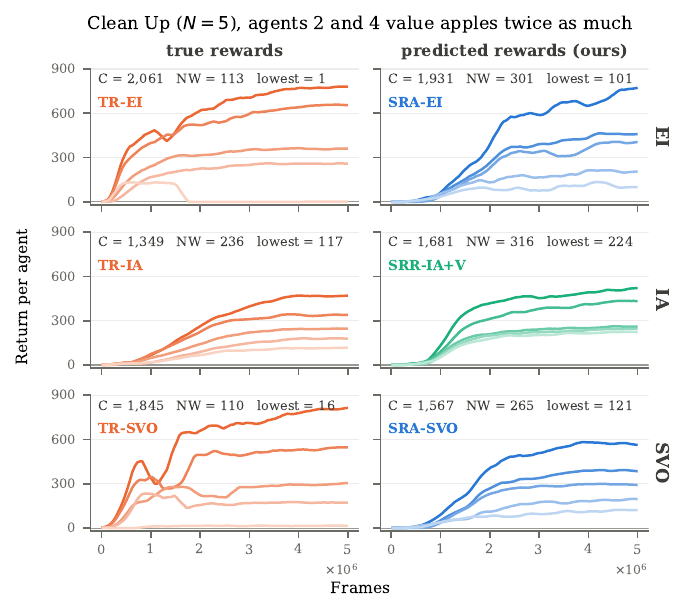}
  \caption{Who receives the return in Clean Up when agents 2 and 4 earn two
  points per apple. Rows are operators and columns the true-reward learner
  and ours, each at the setting selected under shared rewards. Each line is
  one agent's return over training, ranked within each seed from the
  highest to the lowest earner and averaged by rank over the eight held-out
  seeds. Our learners reach a higher Nash welfare and leave the poorest
  agent more than the true-reward learner of the same operator; under
  \alg{SRR-IA+V} the agents eat almost equally many apples, so the two
  agents with the higher value end with about twice the return of the
  others.}
  \label{fig:app-agents-hetero}
\end{figure}

We now deliberately violate the condition under which self-reference can recover another agent's reward. Two of five Clean Up agents value an apple at 2 rather than 1. A self-referenced agent therefore cannot recover the other's true reward from its transition: it assigns the transition the value that it would itself receive. The experiment asks whether self-reference remains useful when it no longer estimates the other's private reward, but only assesses the other's situation from the agent's own perspective. Every learner keeps the setting selected under shared rewards. Figure~\ref{fig:app-agents-hetero} shows every agent's return, with the true-reward learner in the left column, ours in the right, and one row per operator.

\begin{figure*}[tp]
  \centering
  \includegraphics[width=\textwidth]{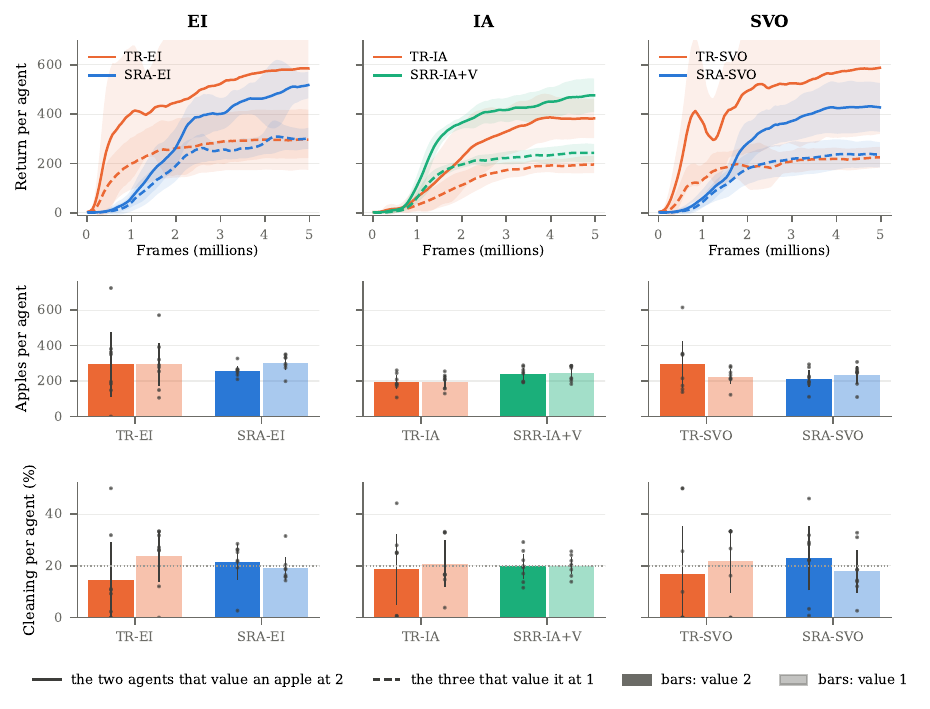}
  \caption{What each learner gives the agents that value apples more, in
  Clean Up with five agents when agents 2 and 4 value an apple at 2. One
  column per operator, comparing the true-reward learner with ours, each at
  the setting selected under shared rewards. Top: return per agent over
  training for the two high-value agents, solid, and the three others,
  dashed, means over the eight held-out seeds with pointwise 95\%
  Student-$t$ intervals. Middle and bottom: final apples eaten and share of
  the cleaning per agent, dark for the high-value agents and light for the
  others, with 20\% an equal share of the cleaning; bars are means with 95\%
  intervals and dots single seeds.}
  \label{fig:app-hetero-groups}
\end{figure*}

In every row the right panel's lines sit closer together than the left
panel's: our learners divide the return more evenly, reaching a higher Nash
welfare and leaving the poorest agent more, 101 against 1 under EI, 224
against 117 under inequity aversion, and 121 against 16 under SVO. The
ranked lines cannot show which agents value apples more, so
Figure~\ref{fig:app-hetero-groups} splits each learner's agents into the two
high-value agents and the three others.

The split shows that no learner divides by value. In every learner,
including the three given the true rewards, the high-value agents eat about
as many apples as the others, and so end with about twice the return. The
true-reward learners lean slightly towards leaving the cleaning to the
low-value agents, whose apples are worth less, but the spread between seeds
is too wide to call it an effect. What separates the learners is
consistency. \alg{SRR-IA+V} gives every agent nearly the same apples and the
same share of the work on every seed, while the true-reward learners swing
widely, and on some seeds a single agent eats more than 700 apples.

The doubled return of the high-value agents is therefore not a consequence of self-reference, since agents that observe the values produce it too. Self-referenced agents cannot see the difference: each assesses every apple at its own value. Here, however, the agents that can see the difference do not act on it either, so the counterfactual assessment costs nothing that the true rewards would have gained. Self-reference stays useful when it is no longer estimating the other's private reward. A real test
of whether judging others by one's own values misleads an agent would need a
preference that responds to the value gap when it is visible, such as
inequity aversion on true rewards with a strong penalty on being ahead,
which the selection for shared rewards did not choose.

\FloatBarrier
\section{Limitations}
\label{app:limitations}

Our conclusions hold within the limits of the study, which we collect here.

\paragraph{The shared reward function.} A self-referenced estimate recovers another agent's reward only when every agent is rewarded for the same events in the same way. When values differ, it becomes a counterfactual assessment: the reward that the agent itself would receive in the other's situation. We test one such case, agents that value an apple differently (Appendix~\ref{app:hetero}). There no learner acts on the difference, so the test cannot show whether self-reference would mislead an agent that otherwise would. Agents whose rewards come from different events, or from events that their transitions do not show, are beyond a model of one's own reward.

\paragraph{One learning algorithm.} Every learner is independent PPO. The
reward placement needs only a reward and the policy-update placement only a
weighted policy gradient, so both should carry over to other actor-critic
methods, and the reward placement to value-based learners, but we have not
tested either.

\paragraph{Selection on few seeds.} Every setting is chosen on two or three
tuning seeds by the mean minus one standard deviation, a rule of thumb
rather than a confidence bound. Its failures are visible: plain PPO with a
recurrent policy, selected this way, earns three apples in Clean Up with the
partial view (Appendix~\ref{app:po-sra}), and one \alg{TR-SVO} setting
produced more on the held-out seeds than the setting selected for it
(Appendix~\ref{app:pick-plane}). Because every number comes from held-out
seeds, these failures add noise to the comparison but do not favour any
learner.

\paragraph{Selected learners, not isolated placements.} The two placements
are compared at their selected settings, which differ in their social
weights and PPO settings as well as in where the estimate enters. The
comparison therefore describes the learners we would deploy rather than
isolating the placement, and the mechanisms we give for it, such as why
the true rewards hinder inequity aversion in Clean Up and as the group
grows, are explanations we have not tested directly.

\paragraph{Estimation in Commons Harvest.} In Commons Harvest the reward
model is a small head on the policy's features rather than a model of the
transition (Appendix~\ref{app:settings}), and the large beam penalty
dominates what it learns. This is the game where our learners fall
furthest short of the true rewards, and a better model of the transition
may narrow the gap.

\paragraph{Welfare as a measure.} The Nash welfare rewards balance, and it
can be raised without cooperation: in Commons Harvest a control that
ignores the others reaches the highest welfare of all
(Appendix~\ref{app:tables}). We therefore read it together with the
collective and the lowest agent return, and never alone.

\paragraph{Cost.} Every agent evaluates its reward model and critic on
every other agent's stream, which costs $\mathcal{O}(N^2)$ forward passes
per step. This is affordable for the group sizes we study, up to twelve
agents, but it grows quickly beyond them.

\FloatBarrier
\section{Extended Related Work}
\label{app:related}

Our work sits where three lines of research meet: social preferences that
help learning agents cooperate, models of other agents built from one's
own, and the information that multi-agent methods assume they can read
from the other agents. We review each, and close with fairness, which we
use to evaluate rather than to train. Table~\ref{tab:access} summarizes
what each family of methods reads from the others.

\paragraph{Sequential social dilemmas and social preferences.}
Sequential social dilemmas extend matrix-game dilemmas to environments in
which cooperation and defection are policies rather than single
actions~\cite{Leibo2017}; common-pool resource games such as Commons
Harvest show how independent learners deplete a shared
resource~\cite{Perolat2017}, and Melting Pot collects many such games into
a benchmark suite~\cite{Leibo2021,Agapiou2022}. Social preferences are the
most direct remedy. Hughes et al.~\cite{Hughes2018} give agents the
inequity aversion of Fehr and Schmidt~\cite{Fehr1999}, computed on traces
of every agent's reward, and show that aversion to being ahead sustains
cleaning in Clean Up while aversion to being behind sustains punishment in
Commons Harvest; their agent is our \alg{TR-IA}. McKee et al.~\cite{McKee2020} give agents social value orientations and study how the diversity of orientations in a population shapes its behaviour. Peysakhovich and
Lerer~\cite{Peysakhovich2018} show that prosocial agents, which add the
others' rewards to their own, find the cooperative equilibrium of stag
hunts more often than selfish ones. Wei et al.~\cite{Wei2026}
combine altruistic and fairness preferences through a constant-elasticity-of-
substitution (CES) utility and apply it to expected future returns, providing
a forward-looking social utility. Their utility still takes other agents'
returns as inputs; we instead replace the unavailable peer-outcome input
with a self-referenced estimate. Further work has learned
reciprocity~\cite{Eccles2019}, evolved intrinsic social motivations~\cite{Wang2019}, and compared moral preferences in the same
dilemmas~\cite{Tennant2023}. Conditional cooperation, as in
approximate tit-for-tat~\cite{Lerer2017}, trains a cooperative policy on
the sum of all agents' rewards and switches to it when the partner
cooperates. These approaches generally rely on other agents' rewards, outcomes, or welfare signals during training. We keep their preferences unchanged and replace
only that input.

\paragraph{Changing the incentives.}
A second line changes the rewards agents receive rather than what they
value. Agents can gift part of their reward~\cite{Lupu2020}, learn to pay
incentives that change how others learn~\cite{Yang2020}, share rewards
along a network~\cite{Yi2022}, learn how much of the others' losses to mix into their own to reduce the price of anarchy~\cite{Gemp2022}, agree on enforceable
contracts~\cite{Haupt2024}, or transfer the smallest amount of reward that
makes cooperation stable~\cite{Willis2024}. These mechanisms alter or transfer the rewards available to agents, rather than infer another agent's outcome from its observed experience.
Vinitsky et al.~\cite{Vinitsky2023} take a different route that is close in
spirit to ours: their agents learn which behaviours a group sanctions by
watching public sanctions, a social signal learned from observation rather
than read from rewards. Social influence~\cite{Jaques2019} also achieves coordination without access to others' rewards by rewarding causal influence over their actions;
its decentralized variant can compute this signal from observed actions using
a learned model of other agents. Unlike our method, however, influence is
the social signal itself rather than a representation of the other's
outcome. Conventions can also emerge among agents with different
preferences without social influence~\cite{Koster2020}.

\paragraph{What cooperative methods read from the others.}
Cooperative multi-agent reinforcement learning usually assumes a team
reward. Value decomposition learns one value per agent that adds
up~\cite{Sunehag2018} or combines monotonically~\cite{Rashid2018} into the
team's value, and centralized training with decentralized execution gives
each agent a critic that sees the others' observations and actions during
training~\cite{Lowe2017,Foerster2018COMA,Yu2022MAPPO}. Counterfactual
baselines~\cite{Foerster2018COMA} and difference
rewards~\cite{Wolpert2001} assign the team reward to individual agents,
and opponent shaping~\cite{Foerster2018LOLA} differentiates through the
other agent's learning step. Our setting shares what centralized training
shares, the observations and actions, but in a general-sum game with no
team reward and no access to the others' rewards, critics, or gradients.
\alg{SRA} resembles a counterfactual baseline in form, since it weights an
agent's update by how the others' transitions went, but it uses no joint
action, no shared reward, and no gradient into another agent. Independent
PPO, which we build on, is a strong baseline in this
family~\cite{deWitt2020,Yu2022MAPPO}. \alg{SRR} is a form of reward
shaping, but not a potential-based one~\cite{Ng1999}: it is meant to
change which policies are optimal, as every social preference does.

\paragraph{Modelling others from one's own experience.}
Agents that model other agents are surveyed by Albrecht and
Stone~\cite{Albrecht2018}. Deep opponent models learn to predict the
others' actions from their observations~\cite{He2016}, and machine theory
of mind learns to predict another agent's behaviour and false beliefs from
a few observations~\cite{Rabinowitz2018}. Closest to our view of empathy
are methods that use the agent's own machinery to understand another. Self
Other-Modeling infers another agent's goal by running the agent's own
policy on the other's observations~\cite{Raileanu2018}. Empathic deep
$Q$-learning evaluates the agent's own $Q$-function from the other's
perspective to avoid harming it~\cite{Bussmann2019}. Sympathy-based agents infer another agent's reward and incorporate it into their own objective~\cite{Senadeera2022}; EMOTE instead uses the learner's own action-value function as a reference for modelling the other agent~\cite{Senadeera2023}. LASE uses perspective taking to
decide how much reward to give away to a co-player~\cite{Kong2024LASE},
and homeostaticcoupling models prosocial behavior through shared or inferred homeostatic states~\cite{Yoshida2025}.. We share their premise that one's
own experience is the natural model of another's, and differ in what we do
with it: we assess the other's transition with a reward model fitted on the agent's own transitions, feed that self-referenced estimate to standard social preferences, and ask where it should enter learning.

\paragraph{Inferring rewards from behaviour.}
Inverse reinforcement learning infers rewards from behavior under assumptions about the other's decision process~\cite{Ng2000}, and has been extended to multi-agent settings~\cite{Natarajan2010,Yu2019MAAIRL}; cooperative inverse reinforcement learning lets a robot learn a human's reward while acting with them~\cite{HadfieldMenell2016}. In cognitive science, inverse planning infers others' goals and desires from their behavior under an assumed planning model~\cite{Baker2009,JaraEttinger2019}, and related models infer higher-level intentions in social interaction~\cite{KleimanWeiner2016}. These approaches infer latent preferences, goals, or intentions from what the other does. Self-reference instead evaluates what happens to the other using the agent's own reward model.

\paragraph{Private rewards and communication.}
Some work keeps rewards private but lets agents exchange other signals.
In suggestion sharing, agents exchange suggested actions instead of
rewards or values~\cite{Jin2025}, and in multi-agent reward prediction a
reward model trained on external, episode-level assessments of the
group's outcome steers decentralized agents towards goals such as
sustainability and equality~\cite{Caftory2026}. Our agents exchange
nothing and receive no evaluation beyond their own reward.

\paragraph{Empathy and social preferences in people.}
Social preferences have long been studied in people, particularly in laboratory games~\cite{Fehr1999,Bolton2000,Charness2002,Fehr2000,Camerer2003} and
across societies~\cite{Henrich2001}. Psychology separates the cognitive
part of empathy, estimating another's state, from its motivational part,
caring about it~\cite{Batson1991,Decety2004,deWaal2008}. The estimate is
explained by mapping the other's situation onto one's own
representations~\cite{Preston2002}, by simulating oneself in the other's
place~\cite{Gordon1986,Goldman2006}, or by inverse
planning~\cite{Baker2009,JaraEttinger2019}; self-reference follows the
simulation route. Simulation has a known cost: people assume that others
share their preferences~\cite{Ross1977} and adjust too little from their
own perspective~\cite{Epley2004}. Our experiment with different values in
Appendix~\ref{app:hetero} is a first test of whether learning agents
inherit that bias. Cooperation across repeated encounters, which our
standing ledger addresses, has a long history of its own, beginning with
the evolution of reciprocal strategies~\cite{Axelrod1981}.

\paragraph{Fairness.}
Fairness has been built into multi-agent learning by centralized
mechanisms~\cite{Grupen2022}, by optimizing welfare functions of the agents' returns~\cite{Jiang2019,Zimmer2021,Siddique2020}, by comparing egoistic, utilitarian, and egalitarian objectives~\cite{Dong2024}, and, in
asymmetric sequential social dilemmas, by correcting social incentives
so that agents that differ are treated fairly rather than
equally~\cite{Demir2026}. All of these need the agents' returns. We evaluate fairness using: the Nash welfare~\cite{Kaneko1979}, which
rewards both production and balance and has strong fairness properties in
allocation~\cite{Caragiannis2019}, and the lowest agent return, the
maximin criterion of Rawls~\cite{Rawls1971}.

\begin{table}[tb]
  \centering
  \caption{What each method uses from the other agents during training
  ($\bullet$: used; $\circ$: in some variants): rewards, observations and
  actions, policies or gradients, and a centralized critic or joint value. At execution
  every method listed acts on the agent's own observation.}
  \label{tab:access}
  \centering
  \footnotesize
  \setlength{\tabcolsep}{4pt}
  \begin{tabular}{@{}lcccc@{}}
    \toprule
    Method & Rewards & Obs./act. & Policies & Critic \\
    \midrule
    Inequity aversion~\cite{Hughes2018} & $\bullet$ & & & \\
    Social value orientation~\cite{McKee2020} & $\bullet$ & & & \\
    Prosocial reward~\cite{Peysakhovich2018} & $\bullet$ & & & \\
    VDN, QMIX~\cite{Sunehag2018,Rashid2018} & $\bullet$ (team) & $\bullet$ & & $\bullet$ \\
    COMA~\cite{Foerster2018COMA} & $\bullet$ (team) & $\bullet$ & & $\bullet$ \\
    MADDPG~\cite{Lowe2017} & & $\bullet$ & $\circ$ & $\bullet$ \\
    LOLA~\cite{Foerster2018LOLA} & $\bullet$ & $\bullet$ & $\bullet$ & \\
    Social influence~\cite{Jaques2019} & & $\bullet$ & $\circ$ & \\
    \midrule
    \alg{SRR}, \alg{SRA} (ours) & & $\bullet$ & & \\
    \bottomrule
  \end{tabular}
\end{table}

The table makes the position of our learners plain. The social-preference
methods need only the others' rewards, which is exactly the channel we
remove. The cooperative and centralized methods read observations and
actions as we do, but also a team reward, a centralized critic, or the
others' policies and gradients. Our learners occupy the one row that reads
observations and actions alone, so they ask for less than centralized
training with decentralized execution and, unlike the social preferences,
for nothing that an agent could not have obtained by watching.


\end{document}